\documentclass[12pt]{article}
\usepackage{amsmath}
\usepackage{graphicx,psfrag,epsf}
\usepackage{enumerate}
\usepackage{natbib}
\usepackage{url} 

\usepackage{amssymb, amsfonts, amscd, xspace, pifont, amsthm}
\usepackage{mathrsfs}
\usepackage{booktabs}
\usepackage{multirow}
\usepackage{epstopdf}
\usepackage{color}
\usepackage[lined,boxed,commentsnumbered]{algorithm2e}
\usepackage{tikz}
\usetikzlibrary{fit,positioning,arrows,automata}

\newcommand{\blind}{0}

\newcommand{\V}[1]{\ensuremath{\boldsymbol{#1}}\xspace}

\newtheorem{thm}{Theorem}[section]
\newtheorem{lemma}{Lemma}[section]

\numberwithin{equation}{section}
\def\twoImages#1#2#3#4#5#6 
{
  \centerline{\hfill\makebox[#2]{#3}\hfill\makebox[#5]{#6}\hfill}
  \centerline{\hfill
    \includegraphics[width=#2]{#1}
    \hfill
    \includegraphics[width=#5]{#4}
    \hfill}
}

\newcommand{\tb}{\textbf}
\DeclareMathOperator*{\argmax}{arg\,max}

\newcommand\norm[1]{\left\lVert#1\right\rVert}

\begin{document}

\def\spacingset#1{\renewcommand{\baselinestretch}%
{#1}\small\normalsize} \spacingset{1}

\if0\blind
{
  \title{Logistic Regression Augmented Community Detection for Network Data with Application in Identifying Autism-Related Gene Pathways}

\author{Yunpeng Zhao,
	Qing Pan, and
	Chengan Du \\
	}
  \maketitle
} \fi

\if1\blind
{
  \bigskip
  \bigskip
  \bigskip
  \begin{center}
    {\LARGE\bf Logistic Regression Augmented Community Detection for Network Data with Application in Identifying Autism-Related Gene Pathways}
\end{center}
  \medskip
} \fi

\bigskip
\begin{abstract}
	When searching for gene pathways leading to specific disease outcomes, additional information on gene characteristics is often available that may facilitate to differentiate genes related to the disease from irrelevant background when connections involving both types of genes are observed and their relationships to the disease are unknown. We propose method to single out irrelevant background genes with the help of auxiliary information through a logistic regression, and cluster relevant genes into cohesive groups using the adjacency matrix. Expectation-maximization algorithm is modified to maximize a joint pseudo-likelihood assuming latent indicators for relevance to the disease and latent group memberships as well as Poisson or multinomial distributed link numbers within and between groups. A robust version allowing arbitrary linkage patterns within the background is further derived. Asymptotic consistency of label assignments under the stochastic blockmodel is proven. Superior performance and robustness in finite samples are observed in simulation studies. The proposed robust method identifies previously missed gene sets underlying autism related neurological diseases using diverse data sources including de novo mutations, gene expressions and protein-protein interactions.
\end{abstract}

\newpage
\spacingset{1.45}

\section{Introduction}
\label{s:intro}

Community detection is a fundamental question in network analysis \citep{Goldenberg2010, NewmanPNAS, Fortunato2010}. Traditional approaches consider the adjacency matrix, whose elements equal one or zero indicating whether there is a connection between two nodes, as the input. Then the nodes are partitioned into cohesive groups, that is, communities, with homogeneous linkage probabilities within and heterogeneous probabilities between the groups. Current community detection methods assume all nodes belong to certain communities of interests. However, this assumption is not always true in real applications. For example, when we are looking for pathways involving genes related to the risk of target disease, connections between candidate genes regardless their involvement in the disease process are collected. Furthermore, whether a gene has impacts on the disease origination and development is usually unknown. Often, information on the characteristics of the nodes/genes can help to differentiate the nodes related to the outcome of interests and the unrelated ones. Novel two-stage models with one joint likelihood are proposed to incorporate the node-specific information which isolate irrelevant nodes from relevant ones and in return improve detection accuracy of communities related to a specific outcome.

Our study is motivated by the problem to discover gene pathways leading to complex diseases in genomic studies. Highly correlated gene expression levels and experimentally verified protein-protein interactions provide useful information on connections between genes. However, not all genes are related to the disease under study. In fact, most genes are ``household" genes with functions to maintain normal metabolic processes within healthy human bodies. Mixing genes and pathways for normal life processes with those leading to the target disease in community detection models will introduce noise to disease-generating pathways which are the true interests of clinicians and biologists. De novo mutations refer to gene mutations that occur for the first time in a family compared to mutations inherited from parents. We believe that discrepancy in the numbers of de novo mutations on the same gene in patients and the number in healthy controls would help differentiate genes related to the disease from those unrelated to the disease, which we call the ``background''. The three kinds of data, gene expression, protein-protein interaction and number of de novo mutations, can be downloaded from different online data consortiums and combined using unique gene names.

The stochastic blockmodel is the most used statistical tool for modeling and detecting communities \citep{Holland83, Snijders&Nowicki1997, Nowicki2001}. We model the relationship between the unobserved indicator whether a gene is related to the target disease or not and gene-specific covariates by logistic regression in the first stage, then cluster disease-related genes into several pathways in the second stage. Both indicators for disease relevance in the first stage and community labels in the second stage are latent variables, and the expectation-maximization algorithm is employed. However, this approach is intractable due to the numerous possible label assignments in the E-step. \citet{AAA} proposed a fast pseudo-likelihood algorithm for fitting blockmodels and we adapt this algorithm in Section \ref{sec:alg} to the joint pseudo-likelihoods incorporating both the logistic regression and the block models. The pseudo-likelihood may also be optimized by other alternative approaches such as the EMM algorithm by \citet{gormley2008mixture}.

Another distinct feature of the proposed method is the extension to the robust community detection allowing heterogeneous linkage probabilities in the background, which relaxes the assumption of homogeneous linkage probability within each group in the stochastic blockmodel. For instance, the background can be a mixture of multiple strongly or weakly connected groups. These groups all belong to the background because they are not related to the target disease, but their linkage rates are not necessarily homogeneous. In Section \ref{sec:robust}, we further develop the model in section 3 to allow for arbitrary linkage patterns within the background. Interestingly, when the linkage probabilities within the background are unspecified, the pseudo-likelihood algorithm can be modified to leave the likelihood of the links in the background out while the classical likelihood approach cannot.

Recently there have been works on community detection which utilize covariates information. These papers use the additional covariates information to improve the accuracy of community detection. Some papers combine a similarity or kernel matrix based on covariates with the adjacency matrix \citep{Rohe2014,Zhang2015,Sarkar2016,Xu2012}. Other papers build likelihoods of linkage probabilities incorporating auxiliary nodal information \citep{Tallberg2004,Yang13,Newman2016,Handcock2007,Hoff2009,gormley2010mixture}. However, none of these works follow the same framework as our method. In short, in our method, the sole reason of using auxiliary information on nodal characteristics is to distinguish the disease related nodes from unrelated ones, then we carry out community detection within the disease-related nodes. On the contrary, in the literature, auxiliary information is used to facilitate partition of all nodes into communities. For example, \citet{Tallberg2004} used covariates to predict the probabilities into each homogeneous community in a Bayesian framework, while we use covariates to predict the probability into the heterogeneous background in a pseudo-likelihood framework.

\section{Methods} \label{sec:method}

We begin by introducing the data structure and notation. A network with $n$ nodes can be represented by
an $n\times n$ adjacency matrix $A=[A_{ij}]$, where


\[
A_{ij}=\begin{cases}
1 & $if there is an edge between $ i $ and $ j, \\
0 & $otherwise$ \\
\end{cases}
\]
In addition to the adjacency matrix $A$, some covariate information on nodes is also available. These covariates are represented by an $n \times P$ matrix $X=[x_{ip}]$, where $x_{ip}$ denotes the value of the $p$th covariate on node $i$.

We model networks with a particular community structure where the network is composed of multiple cohesive communities, together with some \textit{background} nodes. Unlike the usual definition of background set which is diffuse within itself or weakly connected to other parts of the network \citep{Zhao.et.al.2011}, we assume that the probability of a node belonging to the background set depends on its covariates. Suppose there are $K$ communities besides the background set. Let $\V{c}=(c_1,c_2,...,c_n)$ denote the community that each of the $n$ nodes/genes belongs to, thus  $c_i=k$ if nodes $i$ belongs to community $k$, for $k \in \{1,2,...,K\}$, and $c_i=K+1$ if node $i$ is a background gene. Moreover, let $\V{y}=[y_i]$ be a vector indicating whether the node belongs to one of the $K$ communities or the background, i.e. $y_i=1$ if $c_i\leq K$, $y_i=0$ otherwise.

The  network is generated in three steps.
\begin{enumerate}
\item 	The random variable $y_i$ is independent for $i=1,\cdots,n$ and follows a logistic regression
	\begin{align}
	\textnormal{pr}(y_i=1 \mid X)=\frac{e^{\V{x}_i \V{\beta}}}{1+e^{\V{x}_i\V{\beta}}}, \nonumber
	\end{align}
	where $\V{\beta}=(\beta_1,...,\beta_P)^T$ is the coefficients vector, and $\V{x}_i$ is the $i$th row of $X$. Here the logistic model has an intercept, that is, the first column of $X$ is $(1,1,...,1)^T$.

\item 	The probability that a node with $y_i=1$ belongs each of the $K$ communities is given by the independent multinomial distribution with parameter $\V{\pi}=(\pi_1,...,\pi_K)$,
	\begin{align}
	& \textnormal{pr}(c_i=k \mid y_i=1)= \pi_k, \quad (i=1,...,n; k=1,...,K). \nonumber
	\end{align}
	In addition, $c_i=K+1$ if $y_i=0$.

\item 	Conditional on the labels, $A_{ij}$ for $i<j$ are independent Bernoulli variables with
	\begin{align}
	\textnormal{pr}(A_{ij}=1 \mid \V{c})=P_{c_ic_j}, \nonumber
	\end{align}
	where $P$ is a $(K+1)\times (K+1)$ symmetric matrix.
\end{enumerate}
The total number of genes in the $k$th community is $n_k=\sum_{i=1}^n 1(c_i=k)$ and the number of links between the $k$th and $l$th commuity is given by $O_{kl}=\sum_{1\leq i, j\leq n} A_{ij}1(c_i=k,c_j=l)$, where $1(\cdot)$ is the indicator function.  Moreover, let $n_{kl}=n_k n_l$ if $k \neq l$, and $n_{kk}=n_k(n_k-1)$.
Then the joint log-likelihood of $\V{c}$ and $A$ is
\begin{align}\label{block}
\mathcal{L} (\V{\beta},\V{\pi},P ; \V{c},A)= & \sum_{i=1}^n \{ y_i \V{x}_i\V{\beta}- \log(1+e^{\V{x}_i\V{\beta}} )\}+ \sum_{k=1}^K n_k \log \pi_k \nonumber  \\
& + \frac{1}{2}\sum_{1\leq k,l \leq K+1} \left \{ O_{kl} \log P_{kl}+(n_{kl}-O_{kl})\log (1-P_{kl}) \right \}.
\end{align}

\section{Estimating Procedures}\label{sec:alg}
The community labels $\V{c}$ are unobserved in a community detection problem. Furthermore, the E-step of such algorithm requires evaluating all possible label assignments, which makes the algorithm intractable \citep{AAA,Zhaoetal2012}. We adopt the idea of pseudo-likelihood in \cite{AAA} which partitions each row of $A$ into blocks and assumes the independence between rows.

We use the same notation as those in \cite{AAA}. The vector $\V{e}=(e_1,...,e_n)$ denotes an initial blocking vector, where $e_i\in \{1,...,K+1\}$. And $b_{ik}$ denotes the number of edges associated with node $i$ in the $k$th block, that is, $b_{ik}=\sum_{j=1}^n A_{ij} 1(e_j=k) \,\, (i=1,..,n; k=1,...,K+1)$. Let $B=[b_{ik}]_{1\leq i \leq n, l \leq k \leq K+1}$ and $\Lambda=[\lambda_{lk}]_{1\leq l,k \leq K+1}$, where $\lambda_{lk}$ is the expected total number of edges in the $k$-th block for a node $i$ in community $l$, i.e., $c_i=l$. When $n$ is large, $b_{ik}$ can be approximated by a Poisson distribution given $c_i$, and the dependence of $B$ between different rows is weak. Assuming $b_{ik}$ are independence for $i=1,\cdots,n$ and $k=1,\cdots,K+1$  and using the Poisson approximation, the log-pseudolikelihood of $\V{c}$ and $B$ (up to a constant) is
\begin{align}
\sum_{i=1}^n \{ y_i \V{x}_i\V{\beta}- \log(1+e^{\V{x}_i\V{\beta}} )\}+ \sum_{k=1}^K n_k \log \pi_k  +\sum_{i=1}^n \sum_{l=1}^{K+1} 1(c_i=l) \left ( -\mu_l  +\sum_{k=1}^{K+1} b_{ik} \log \lambda_{lk} \right ), \nonumber
\end{align}
where $\mu_{l}=\sum_{k} \lambda_{lk} \,\, (l=1,...,K+1)$. And the log-likelihood for the marginal distribution of $B$ (up to a constant) is
\begin{align}\label{Poisson}
\mathcal{L}_{\mbox{Poisson}} (\V{\beta},\V{\pi},\Lambda ; B)= & \sum_{i=1}^n \log \left \{  \sum_{l=1}^K \frac{e^{\V{x}_i \V{\beta} }}{1+e^{\V{x}_i \V{\beta} }}\pi_l e^{-\mu_l} \left ( \prod_{k=1}^{K+1}  \lambda_{lk}^{b_{ik}} \right ) \right . \nonumber \\
& + \left . \frac{1}{1+e^{\V{x}_i \V{\beta} }} e^{-\mu_{K+1}} \left ( \prod_{k=1}^{K+1}  \lambda_{K+1,k}^{b_{ik}} \right ) \right \}.
\end{align}
Given initial labels $\V{e}$, equation \eqref{Poisson} can be maximized by a standard expectation-maximization algorithm. The details of the E-step and M-step are given in Algorithm 1.

\textbf{Algorithm 1:} (The expectation-maximization algorithm under Poisson distribution)
\begin{itemize}
	\item E-step: Let $\hat{\V{\beta}}, \hat{\V{\pi}}$ and $\hat{\Lambda}$ be the estimates at the current iteration, and $\hat{\mu}_l=\sum_{k} \hat{\lambda}_{lk}\,\, (l=1,...,K+1)$. The posterior probability of label assignment is
	\begin{align}
	z_{il} & =\textnormal{pr}(c_i=l \mid B ) \nonumber \\
	& =\frac{\frac{e^{\V{x}_i\hat{\V{\beta} }}}{1+e^{\V{x}_i \hat{\V{\beta} }}}\hat{\pi}_l e^{-\hat{\mu}_l} \left ( \prod_{k=1}^{K+1}  \hat{\lambda}_{lk}^{b_{ik}} \right )}{\sum_{l=1}^K \frac{e^{\V{x}_i\hat{\V{\beta}} }}{1+e^{\V{x}_i\hat{\V{\beta} }}}\hat{\pi}_l e^{-\hat{\mu}_l} \left ( \prod_{k=1}^{K+1}  \hat{\lambda}_{lk}^{b_{ik}} \right )  + \frac{1}{1+e^{\V{x}_i\hat{\V{\beta}} }} e^{-\hat{\mu}_{K+1}} \left ( \prod_{k=1}^{K+1}  \hat{\lambda}_{K+1,k}^{b_{ik}} \right ) } \nonumber \\
	& \quad \quad \quad \quad \quad \quad \quad \quad \quad \quad \quad \quad \quad \quad \quad  \quad \quad \quad \quad \quad \quad  \quad \quad (i=1,...,n; l=1,...,K), \nonumber \\
	z_{i,K+1}& =\textnormal{pr}(c_i=K+1 \mid B ) \nonumber \\
	&  =\frac{\frac{1}{1+e^{\V{x}_i\hat{\V{\beta}} }} e^{-\hat{\mu}_{K+1}} \left ( \prod_{k=1}^{K+1}  \hat{\lambda}_{K+1,k}^{b_{ik}} \right )}{\sum_{l=1}^K \frac{e^{\V{x}_i\hat{\V{\beta}} }}{1+e^{\V{x}_i\hat{\V{\beta} }}}\hat{\pi}_l e^{-\hat{\mu}_l} \left ( \prod_{k=1}^{K+1}  \hat{\lambda}_{lk}^{b_{ik}} \right )  + \frac{1}{1+e^{\V{x}_i\hat{\V{\beta}} }} e^{-\hat{\mu}_{K+1}} \left ( \prod_{k=1}^{K+1}  \hat{\lambda}_{K+1,k}^{b_{ik}} \right ) } \nonumber \\
	& \quad \quad \quad \quad \quad \quad \quad \quad \quad \quad \quad \quad \quad \quad \quad  \quad \quad \quad \quad \quad \quad  \quad \quad (i=1,...,n). \nonumber
	\end{align}
	
	\item M-step: Given $z_{il} \,\, (i=1,...n; l=1,...,K+1)$, $\hat{\pi}$ and $\hat{\Lambda}$ can be updated by the following formulae,
	\begin{align}
	\hat{\pi}_{l} & = \frac{\sum_{i} z_{il}}{\sum_{i} \sum_{l=1}^K z_{il} } \quad (l=1,...,K), \nonumber \\
	\hat{\lambda}_{lk} & =\frac{\sum_i z_{il} b_{ik}}{\sum_i z_{il}} \quad(l=1,...,K+1; k=1,...,K+1) . \nonumber
	\end{align}
	$\hat{\V{\beta}}$ can be updated by solving the logistic regression,
	\begin{align}
	\hat{\V{\beta}} & = \argmax_{\V{\beta}} \sum_{i=1}^n \left \{  \left (\sum_{l=1}^K z_{il} \right ) \V{x}_i\V{\beta}- \log(1+e^{\V{x}_i\V{\beta}} )  \right \}. \nonumber
	\end{align}
	Note $\sum_{l=1}^Kz_{il}$ is the sum of the estimated conditional probabilities of gene $i$ belonging to one of the $K$ communities.
\end{itemize}

Once the expectation-maximization algorithm converges, we can update the labels $\V{e}$ by $e_i=\argmax_{1\leq l\leq K+1} z_{il}$. We repeat this procedure several times until $\V{e}$ becomes stable.

\cite{AAA} also introduced a pseudo-likelihood conditional on the node degrees. We generalize this conditional pseudo-likelihood to our scenario. Denote the node degree by $d_i=\sum_{k} b_{ik} \,\, (i=1,...,n)$. Then $(b_{i1},...,b_{i,K+1})$ follows multinomial distribution conditional on label $\V{c}$ and $d_i$. The multinomial log pseudo-likelihood (up to a constant) is
\begin{align}\label{Multinomial}
\mathcal{L}_{\mbox{Multinomial}} (\V{\beta},\V{\pi},\Theta ; B)= & \sum_{i=1}^n \log \left \{  \sum_{l=1}^K \frac{e^{\V{x}_i \V{\beta} }}{1+e^{\V{x}_i \V{\beta} }}\pi_l\left ( \prod_{k=1}^{K+1}  \theta_{lk}^{b_{ik}} \right ) \right .  \\
& + \left . \frac{1}{1+e^{\V{x}_i \V{\beta} }}  \left ( \prod_{k=1}^{K+1}  \theta_{K+1,k}^{b_{ik}} \right ) \right \}, \nonumber
\end{align}
where $\Theta=[\theta_{lk}]$ $(l=1,...,K+1;k=1,...,K+1)$ is the parameter in the multimomial distribution satisfying $\sum_{k=1}^{K+1}\theta_{lk}=1 (l=1,...,K+1)$.

The algorithm is similar to that for the Poisson pseudo-likelihood. We give the details of the expectation-maximization algorithm under the multinomial distribution in Algorithm 2.

\textbf{Algorithm 2:} (The expectation-maximization algorithm under multinomial distribution)
\begin{itemize}
	\item E-step: Based on current estimates $\hat{\V{\beta}}, \hat{\V{\pi}}$ and $\hat{\Theta}$, the posterior probability of label assignment is
	\begin{align}
	& z_{il} =\frac{\frac{e^{\V{x}_i\hat{\V{\beta} }}}{1+e^{\V{x}_i \hat{\V{\beta} }}}\hat{\pi}_l \left ( \prod_{k=1}^{K+1}  \hat{\theta}_{lk}^{b_{ik}} \right )}{\sum_{l=1}^K \frac{e^{\V{x}_i\hat{\V{\beta}} }}{1+e^{\V{x}_i\hat{\V{\beta} }}}\hat{\pi}_l  \left ( \prod_{k=1}^{K+1}  \hat{\theta}_{lk}^{b_{ik}} \right )  + \frac{1}{1+e^{\V{x}_i\hat{\V{\beta}} }}  \left ( \prod_{k=1}^{K+1}  \hat{\theta}_{K+1,k}^{b_{ik}} \right ) } \quad (i=1,...,n; l=1,...,K), \nonumber \\
	& z_{i,K+1}  =\frac{\frac{1}{1+e^{\V{x}_i\hat{\V{\beta}} }}  \left ( \prod_{k=1}^{K+1}  \hat{\theta}_{K+1,k}^{b_{ik}} \right )}{\sum_{l=1}^K \frac{e^{\V{x}_i\hat{\V{\beta}} }}{1+e^{\V{x}_i\hat{\V{\beta} }}}\hat{\pi}_l  \left ( \prod_{k=1}^{K+1}  \hat{\theta}_{lk}^{b_{ik}} \right )  + \frac{1}{1+e^{\V{x}_i\hat{\V{\beta}} }}  \left ( \prod_{k=1}^{K+1}  \hat{\theta}_{K+1,k}^{b_{ik}} \right ) }  \quad (i=1,...,n). \nonumber
	\end{align}
	
	\item M-step: Given $z_{il} \,\, (i=1,...n; l=1,...,K+1)$, $\hat{\pi}$, $\hat{\Theta}$ and $\hat{\V{\beta}}$ can be updated by
	\begin{align}
	\hat{\pi}_{l} & = \frac{\sum_{i} z_{il}}{\sum_{i} \sum_{l=1}^K z_{il} } \quad (l=1,...,K), \nonumber \\
	\hat{\theta}_{lk} & =\frac{\sum_i z_{il} b_{ik}}{\sum_i z_{il} d_i} \quad(l=1,...,K+1; k=1,...,K+1),  \nonumber \\
	\hat{\V{\beta}} & = \argmax_{\V{\beta}} \sum_{i=1}^n \left \{  \left (\sum_{l=1}^K z_{il} \right ) \V{x}_i\V{\beta}- \log(1+e^{\V{x}_i\V{\beta}} )  \right \}. \nonumber
	\end{align}
\end{itemize}

\section{Robust Community Detection }\label{sec:robust}
So far we assume that all the disease-related communities and the background satisfy the stochastic blockmodel assumption. In this section, we propose a new pseudo-likelihood method that allows for arbitrary structure in the background, for example, a mixture of strongly and weakly connected
groups, or nodes with high degree variations. In other words, we keep the stochastic blockmodel assumption in the disease-related communities, but make no assumption on the structure within the background. A network with the heterogeneous background is generated in three steps, of which the first two steps are identical to the first two steps in Section 2. The last step has been modified as follows.

\textsc{Step} 3$^*$:
Conditional on the labels, when $k \leq K$ or $l \leq K$, $A_{ij}$ for $i<j$ are independent Bernoulli variables with
\begin{align}
\textnormal{pr}(A_{ij}=1 \mid c_i=k,c_j=l)=P_{k l}.\nonumber
\end{align}
The link probabilities within the background set, i.e., when $k=K+1$ and $l=K+1$, are not specified.

The joint likelihood (\ref{block}) cannot be used as the criteria to estimate c in this situation because it is maximized when all nodes belong to group $K+1$. By contrast, the pseudo-likelihood method introduced in Section \ref{sec:alg} can be extended to this new scenario. Recall the setup in Section \ref{sec:alg}. Let $\V{e}=(e_1,...,e_n)$ be an initial blocking vector. And $b_{ik}$ denotes the number of edges associated with node $i$ in the $k$th block $(i=1,..,n; k=1,...,K+1)$. When $\V{e}$ is a reasonable initial vector, $b_{ik}$ can be approximated by a mixture of Poisson distributions as before when $k=1,...,K$. However, when $k=K+1$, the distribution of $b_{ik}$ is unknown since the link probabilities within the background are unspecified. Therefore, we exclude this part of unreliable information, and propose the following pseudo-likelihood for robust community detection,
\begin{align}\label{Robust}
\mathcal{L}_{\mbox{Robust}} (\V{\beta},\V{\pi},\Lambda ; B, \V{c})= & \sum_{i=1}^n \{ y_i \V{x}_i\V{\beta}- \log(1+e^{\V{x}_i\V{\beta}} )\}+ \sum_{k=1}^K n_k \log \pi_k \nonumber \\
& +\sum_{i=1}^n \sum_{l=1}^{K+1} 1(c_i=l) \left ( -\mu_l  +\sum_{k=1}^{K} b_{ik} \log \lambda_{lk} \right ),
\end{align}
where $\mu_l= \sum_{k=1}^K \lambda_{lk}\,\, (k=1,..,K)$.

Notice equation \eqref{Robust} is indeed a valid likelihood function conditional on $\V{e}$ because the blocking vector $\V{e}$ and the community labeling vector $\V{c}$ are treated differently in our algorithm. The proof of its identifiability is given in the supplementary material. The blocking vector $\V{e}$ partitions the columns of $A$ into $K+1$ blocks and $b_{ik}$ is the sum in the $i$th row $k$th block. Likelihood \eqref{Robust} does not include $B_{\cdot,K+1}$ -- the last column of $B$ since the Poisson approximation are inappropriate. But this does not affect the range of $c_i$, which is still $\{1,...,K+1\}$. Community detection based on \eqref{Robust} can be viewed as a classic clustering problem on $B$. We need to assign a label from 1 to $K+1$ to each row data point, i.e., each $B_{i\cdot}$, which contains $K+1$ features. But we only use the first $K$ features since the last one is not reliable. Then we update the labelling of each gene in the columns and iterate several times. In each iteration, because the groups in the columns and the grouping results in the rows are considered separately, dropping one column of noise will not result in all genes (rows) falling into the $(K+1)$th group. After each iteration of the outer loop updating $e_i$, genes with similar linkage probabilities with the first $K$ groups are classified into the same group with higher and higher accuracy.

The algorithm is therefore similar to Algorithm 1 and given in the following.

\textbf{Algorithm 3:} (The expectation-maximization algorithm for robust community detection)
\begin{itemize}
	\item E-step: Let $\hat{\V{\beta}}, \hat{\V{\pi}}$ and $\hat{\Lambda}$ be the estimates at the current iteration, and $\hat{\mu}_l=\sum_{k=1}^K \hat{\lambda}_{lk}\,\, (l=1,...,K+1)$. The posterior probability of label assignment is
	\begin{align}
	z_{il} & =\textnormal{pr}(c_i=l \mid B ) \nonumber \\
	& =\frac{\frac{e^{\V{x}_i\hat{\V{\beta} }}}{1+e^{\V{x}_i \hat{\V{\beta} }}}\hat{\pi}_l e^{-\hat{\mu}_l} \left ( \prod_{k=1}^{K}  \hat{\lambda}_{lk}^{b_{ik}} \right )}{\sum_{l=1}^K \frac{e^{\V{x}_i\hat{\V{\beta}} }}{1+e^{\V{x}_i\hat{\V{\beta} }}}\hat{\pi}_l e^{-\hat{\mu}_l} \left ( \prod_{k=1}^{K}  \hat{\lambda}_{lk}^{b_{ik}} \right )  + \frac{1}{1+e^{\V{x}_i\hat{\V{\beta}} }} e^{-\hat{\mu}_{K+1}} \left ( \prod_{k=1}^{K}  \hat{\lambda}_{K+1,k}^{b_{ik}} \right ) } \nonumber \\
	& \quad \quad \quad \quad \quad \quad \quad \quad \quad \quad \quad \quad \quad \quad \quad  \quad \quad \quad \quad \quad \quad  \quad \quad (i=1,...,n; l=1,...,K), \nonumber \\
	z_{i,K+1}& =\textnormal{pr}(c_i=K+1 \mid B ) \nonumber \\
	&  =\frac{\frac{1}{1+e^{\V{x}_i\hat{\V{\beta}} }} e^{-\hat{\mu}_{K+1}} \left ( \prod_{k=1}^{K}  \hat{\lambda}_{K+1,k}^{b_{ik}} \right )}{\sum_{l=1}^K \frac{e^{\V{x}_i\hat{\V{\beta}} }}{1+e^{\V{x}_i\hat{\V{\beta} }}}\hat{\pi}_l e^{-\hat{\mu}_l} \left ( \prod_{k=1}^{K}  \hat{\lambda}_{lk}^{b_{ik}} \right )  + \frac{1}{1+e^{\V{x}_i\hat{\V{\beta}} }} e^{-\hat{\mu}_{K+1}} \left ( \prod_{k=1}^{K}  \hat{\lambda}_{K+1,k}^{b_{ik}} \right ) } \nonumber \\
	& \quad \quad \quad \quad \quad \quad \quad \quad \quad \quad \quad \quad \quad \quad \quad  \quad \quad \quad \quad \quad \quad  \quad \quad (i=1,...,n). \nonumber
	\end{align}
	
	\item M-step: Given $z_{il} \,\, (i=1,...n; l=1,...,K+1)$, $\hat{\pi}$, $\hat{\Lambda}$ and $\hat{\V{\beta}}$ can be updated by,
	\begin{align}
	\hat{\pi}_{l} & = \frac{\sum_{i} z_{il}}{\sum_{i} \sum_{l=1}^K z_{il} } \quad (l=1,...,K), \nonumber \\
	\hat{\lambda}_{lk} & =\frac{\sum_i z_{il} b_{ik}}{\sum_i z_{il}} \quad(l=1,...,K+1; k=1,...,K) , \nonumber \\
	\hat{\V{\beta}} & = \argmax_{\V{\beta}} \sum_{i=1}^n \left \{  \left (\sum_{l=1}^K z_{il} \right ) \V{x}_i\V{\beta}- \log(1+e^{\V{x}_i\V{\beta}} )  \right \}. \nonumber
	\end{align}
\end{itemize}
As before, once the expectation-maximization algorithm converges, $\V{e}$ is updated by $e_i= \\ \argmax_{1\leq l\leq K+1} z_{il}$. We repeat this procedure until $\V{e}$ becomes stable.

We do not consider robust community detection using multinomial approximation because the condition $\sum_{k=1}^{K+1}\theta_{lk}=1 (l=1,...,K+1)$ is invalid if the last column is removed.

\section{Asymptotic Properties}\label{sec:proof}

In this section we study the consistency under stochastic blockmodels. Equation \eqref{Multinomial} has slightly simpler form and theoretical derivations than \eqref{Poisson}. The theoretical analysis in this section will focus on the multinomial pseudo-likelihood.

We begin with the setup, which closely follow \cite{AAA}. The true community labels $\V{c}$ are the parameters of interests, where $\pi_k=1/n\sum\limits_{i}1(c_i=k) \,\, (k=1,2)$. We focus on the case of directed blockmodel. A coupling technique can be used to extend the result to the undirected case analogous to that in \cite{AAA}. Consider the edge matrix
\begin{align*}
P=\frac{1}{n}\left(
\begin{array}{cc}
a_1 & b  \\
b & a_2  \\
\end{array}
\right)=\frac{b}{n}\left(
\begin{array}{cc}
\rho_1 & 1  \\
1 & \rho_2  \\
\end{array}
\right),
\end{align*}
where $\rho_k={a_k}/{b}$.
Here $\rho_1$ and $\rho_2$ remain constant, while $b$ can scale with $n$. The directed blockmodel assumes that all the entries in the adjacency matrix are independent Bernoulli variables without forcing $P$ to be symmetric, that is, $A_{ij} \sim \mbox{Bernoulli}(P_{c_ic_j}) \,\, (i=1,...,n;j=1,...,n)$. For simplicity,  a univariate covariate $\V{x}$ taking values in $(1/n,2/n,...,1)$ is assumed.

We illustrate the consistency of one-step expectation-maximization of the multinomial pseudo-likelihood. Starting from some initial labels $\V{e}$
and initial estimates $\hat{b}$,  $\hat{\rho}_1, \hat{\rho}_2$ of the parameters $b$, $\rho_1$ and $\rho_2$, the initial estimates of $\hat{\beta}_0$ and $\hat{\beta}_1$ are obtained from the logistic regression, that is,
\begin{align*}
(\hat{\beta}_0, \hat{\beta}_1) & = \argmax_{\beta_0,\beta_1} \sum_{i=1}^n \left \{   y_i(\beta_0+x_i\beta_1)- \log(1+e^{\beta_0+x_i\beta_1} )  \right \}.
\end{align*}
Define
\begin{align*}
\hat{\pi}_{i1} & = \frac{e^{\hat{\beta}_0+x_i\hat{\beta}_1} }{1+e^{ \hat{\beta}_0+x_i\hat{\beta}_1 }} \quad (i=1,...,n), \\
\hat{\pi}_{i2} & = \frac{1}{1+e^{ \hat{\beta}_0+x_i\hat{\beta}_1 }}   \quad (i=1,...,n).
\end{align*}

Let
\begin{align*}
\hat{P}=\frac{\hat{b}}{n}\left(
\begin{array}{cc}
\hat{\rho}_1 & 1  \\
1 & \hat{\rho}_2  \\
\end{array}
\right),
\end{align*}
and $R$ be the 2 by 2 matrix with entries $\{R_{ka}\}$ given by $R_{ka}=(1/n)\sum\limits_{i=1}^n 1 (e_i=k,c_i=a)$. The initial estimates $\hat{\Theta}$ is obtained by row normalization of $\hat{\Lambda}=[n R \hat{P}]^T $, that is,
\begin{align*}
\hat{\Theta}=\left(
\begin{array}{cc}
\frac{\hat{\lambda}_{11}}{\hat{\lambda}_{11}+\hat{\lambda}_{12}} & \frac{\hat{\lambda}_{12}}{\hat{\lambda}_{11}+\hat{\lambda}_{12}}  \\
\frac{\hat{\lambda}_{21}}{\hat{\lambda}_{21}+\hat{\lambda}_{22}} & \frac{\hat{\lambda}_{22}}{\hat{\lambda}_{21}+\hat{\lambda}_{22}}  \\
\end{array}
\right).
\end{align*}

With the notation defined above, the output of one-step expectation-maximization is
\begin{align*}
\hat{c}_i(\V{e})=\argmax_{k \in \{1,2\}} \left (\log \hat{\pi}_{ik}+\sum_{l=1}^2 b_{il}\log \hat{\theta}_{kl} \right ) \quad (i=1,...,n).
\end{align*}

We use the mis-classification error rate \citep{Choietal2011,Zhaoetal2012,AAA} to measure the performance of $\hat{c}_i$. That is, define
\begin{align*}
M_n(\V{e})=\min_{\phi \in \{ (1 2), (2 1) \} }\frac{1}{n} \sum_{i=1}^n 1 \{ \hat{c}_i(\V{e}) \neq \phi(c_i) \},
\end{align*}
where $\{(1 2),(2 1)\}$ is the set of permutations of $\{1,2\}$. In this definition we consider all $\phi$ values that are permutations of each other because they result in the same community structure.

Consider the class of initial labels that correctly classify the node $i$ as a member of community $k$. The fraction of such nodes among all nodes belonging to community $k$, $\gamma_k$, is formally given by
\begin{align*}
\mathcal{E}=\{\V{e}: \sum\limits_{i}1(e_i=k,c_i=k)=\gamma_k n_k, k=1,2\},
\end{align*}
where $n_k=\sum\limits_{i}1(c_i=k)$ is the size of community $k$.

An extra condition is introduced to avoid perfect separation of $\V{e}$ in the logistic fit. 
We define the following class
\begin{align*}
\mathcal{F}=\{\V{e}: \sum_{i=\hat{n}_2+1}^n 1(e_i=1 )\leq \hat{n}_1 \tilde{\gamma}_1, \sum_{i=1}^{\hat{n}_1} 1(e_i=1 )\leq \hat{n}_1 \tilde{\gamma}_2   \},
\end{align*}
where $\hat{n}_k=\sum\limits_{i}1(e_i=k)$ is the size of initial estimate of community $k$.

The uniform consistency of $\hat{c}_i$ within the class $\mathcal{E} \cap \mathcal{F}$ is established by the following theorem.

\begin{thm}[Main result]
	\label{mainThm}
	Assume $\gamma_1, \gamma_2 \neq 1/2$ and $0<\tilde{\gamma}_1, \tilde{\gamma}_2<1$. Then under some regularity condition, with sufficiently large $\hat{\rho}_1$, $\hat{\rho}_2$ and $b \rightarrow \infty$, for any $\epsilon$,
	\begin{align*}
	\textnormal{pr} \left [\sup_{\V{e}\in \mathcal{E} \cap \mathcal{F} } M_n(\V{e})>\epsilon \right ] \rightarrow 0, \quad \mbox{as} \,\, n \rightarrow \infty.
	\end{align*}
\end{thm}
The details of the regularity condition and the proof is given in the supplementary material.

The proof of the main theorem depends on a key fact that the log ratio of the estimated probabilities $\hat{\pi}_{i1}$ and $\hat{\pi}_{i2}$ has a uniform bound independent with $n$, for $\V{e} \in \mathcal{F} \cap \mathcal{E}$. This is summarized in the following lemma.

\begin{lemma}
	\label{lemma1}
	Assume $0<\tilde{\gamma}_1, \tilde{\gamma}_2<1$. Then if $\V{e} \in \mathcal{F} \cap \mathcal{E}$, there exist $M$ such that for sufficiently large $n$,
	\begin{align*}
	\left | \log \frac{\hat{\pi}_{i1}}{\hat{\pi}_{i2}} \right |<M,
	\end{align*}
	where $M$ is independent with $n$.
\end{lemma}
The proof is given in the supplementary material.

\section{Simulations} \label{sec:sim}

We first examine the performance of the proposed methods under standard stochastic blockmodel. Each network contains $n=500$ nodes and each setup is repeated 500 times. There are three groups including two disease-related communities and one disease-irrelevant background set. The probability a gene is related to the disease follows a logistic regression with $\mbox{logit} \mbox{ pr}(y_i=1 \mid x_i) = 4 x_i+\beta_0$. Here $y_i$ is the indicator for the $i$th node belonging to a disease-related community and covariate $x_i \sim U(-1,1)$. And $\beta_0=-1,0,1$ correspond to the percentages background $62\%$, $50\%$ and  $38\%$, respectively. Nodes with $y_i=1$ are assigned to two non-overlapping communities with equal probabilities $\pi_1=\pi_2=1/2$.  Pairs within the background, as well as pairs composed of one node in the background and the other node in a disease-related community are linked with probability $0$$\cdot$$1$. The linkage probability between the two non-background communities is 0$\cdot$05, while the linkage probability for pairs within the same community ranges from 0$\cdot$15 to 0$\cdot$25.

\begin{table}
	\def~{\hphantom{0}}
	\caption{Comparison of average adjusted rand index (ARI) $\times 100$ under stochastic blockmodels.
		Numbers within parentheses are empirical standard deviations of ARI $\times 100$. }

   \begin{center}
   		\scalebox{0.7}{
	\begin{tabular}{lccccccc} \\
		\hline \hline
		&\multicolumn{3}{c}{With Logistic Models}&&\multicolumn{3}{c}{Without Logistic Models} \\
		\cline{2-4}  \cline{6-8}
		$p_{11}$  &Poisson & Multinomial & Robust && Poisson & Multinomial & Robust \\
		\hline
		\multicolumn{8}{l}{$62\%$ Background Nodels} \\
		15 & 58 (12) & 57 (13) & 59 (12) && 15 (7) & 15 (8) & 15 (8) \\
		16 & 66 (8) & 66 (9) & 67 (8) && 23 (11) & 24 (11) & 23 (11)   \\
		17 & 72 (7) & 72 (6) & 73 (7) && 34 (13) & 33 (13) & 33 (13) \\
		18 & 77 (5) & 76 (5) & 77 (5) && 48 (14) & 45 (13) & 46 (15) \\
		19 & 81 (5) & 80 (5) & 81 (4) && 61 (11) & 55 (13) & 60 (11) \\
		20 & 85 (4) & 83 (4) & 85 (4) && 70 (8) & 66 (9) & 70 (9) \\
		21 & 88 (3) & 86 (4) & 88 (3) && 78 (6) & 73 (7) & 78 (7) \\
		22 & 91 (3) & 88 (3) & 91 (3) && 83 (4) & 79 (6) & 83 (5) \\
		23 & 93 (3) & 90 (3) & 93 (3) && 87 (4) & 83 (5) & 87 (4) \\
		24 & 94 (2) & 92 (3) & 94 (2) && 90 (3) & 86 (4) & 90 (3) \\
		25 & 96 (2) & 93 (2) & 96 (2) && 93 (3) & 89 (3) & 93 (3) \\
		\hline
		\multicolumn{8}{l}{$50\%$ Background Nodels} \\
		15	&	74	(5)	&	74	(5)	&	74	(5)	&&	44	(10)	&	44	(10)	&	43	(11)	\\
		16	&	78	(4)	&	78	(4)	&	79	(4)	&&	56	(8)	&	56	(8)	&	55	(10)	\\
		17	&	82	(4)	&	82	(4)	&	82	(4)	&&	66	(6)	&	64	(7)	&	66	(7)	\\
		18	&	86	(3)	&	85	(4)	&	86	(3)	&&	74	(6)	&	72	(6)	&	74	(6)	\\
		19	&	89	(3)	&	88	(3)	&	89	(3)	&&	80	(5)	&	78	(5)	&	80	(5)	\\
		20	&	91	(3)	&	90	(3)	&	92	(3)	&&	86	(4)	&	82	(4)	&	86	(4)	\\
		21	&	94	(2)	&	92	(3)	&	94	(2)	&&	89	(3)	&	86	(4)	&	89	(3)	\\
		22	&	95	(2)	&	93	(2)	&	95	(2)	&&	92	(3)	&	89	(3)	&	92	(3)	\\
		23	&	96	(2)	&	95	(2)	&	97	(2)	&&	94	(2)	&	91	(3)	&	94	(2)	\\
		24	&	98	(1)	&	96	(2)	&	97	(1)	&&	96	(2)	&	93	(3)	&	96	(2)	\\
		25	&	98	(1)	&	96	(2)	&	98	(1)	&&	97	(1)	&	94	(2)	&	97	(1)	\\
		\hline
		\multicolumn{8}{l}{$38\%$ Background Nodels} \\
		15	&	82	(4)	&	82	(4)	&	82	(3)	&&	67	(6)	&	67	(6)	&	67	(6)	\\
		16	&	86	(3)	&	86	(3)	&	86	(3)	&&	74	(5)	&	74	(5)	&	74	(5)	\\
		17	&	89	(3)	&	88	(3)	&	89	(3)	&&	81	(4)	&	80	(4)	&	80	(4)	\\
		18	&	91	(3)	&	91	(3)	&	91	(3)	&&	85	(3)	&	84	(4)	&	85	(3)	\\
		19	&	94	(2)	&	92	(2)	&	94	(2)	&&	89	(3)	&	87	(3)	&	89	(3)	\\
		20	&	96	(2)	&	94	(2)	&	96	(2)	&&	92	(2)	&	90	(3)	&	92	(2)	\\
		21	&	97	(1)	&	95	(2)	&	97	(1)	&&	95	(2)	&	92	(2)	&	95	(2)	\\
		22	&	98	(1)	&	96	(2)	&	98	(1)	&&	96	(2)	&	94	(2)	&	96	(2)	\\
		23	&	98	(1)	&	97	(1)	&	98	(1)	&&	97	(1)	&	95	(2)	&	97	(1)	\\
		24	&	99	(1)	&	97	(1)	&	99	(1)	&&	98	(1)	&	96	(2)	&	98	(1)	\\
		25	&	99	(1)	&	98	(1)	&	99	(1)	&&	99	(1)	&	97	(2)	&	99	(1)	\\
		\hline \hline
		
	\end{tabular}
}
    \end{center} 

	\label{table1}
	
\end{table} 

Table \ref{table1} compares the performance of three models - the pseudo-likelihood methods with Poisson and multinomial approximation introduced in Section \ref{sec:alg} as well as the robust community detection method introduced in Section \ref{sec:robust}. For each model, we further compare the two versions where auxiliary nodal information, i.e, logistic regression, is either used or unused. The community detection accuracy is measured by the adjusted rand index (\textsc{ari}) \citep{Vinh10}, a widely-used measure for comparing two partitions. The value of the index is 0 for two independent partitions, and higher values indicate better agreement. The performance of all methods improves as the linkage probability within disease-related community increases, or as the percentage of background nodes decreases. More importantly, the proposed method incorporating auxiliary information through logistic regression always outperforms its counterpart without logistic regression. Moreover, the robust method gives the same performance as the Poisson pseudo-likelihood which suggests the robust method does not lose discriminatory accuracy when data follow standard stochastic block models.  On the other hand, the algorithm fitting multinomial distributions performs slightly worse than the other two methods. Rigorously speaking, the multinomial pseudo-likelihood is an approximation to the degree-corrected blockmodel, which is a generalization of standard blockmodel by allowing more variation on degrees \citep{Zhaoetal2012,Karrer10,AAA}. Therefore, the finite sample performance of multinomial pseudo-likelihood has slightly lower ARI on average since it fits a more complicated model.

Next we consider the setup with heterogeneous background nodes. For any node $i$ in background, we generate $u_i$ from $U(0, 0$$\cdot$$2)$. The linkage probability between a background node $i$ and a disease-related node is $u_i$. For two background nodes $i$ and $j$, the linkage probability is $\sqrt{u_iu_j}$. The rest of the model setups such as the generation mechanism of communities labels, the linkage probabilities within/between communities and linkage probabilities between a community and the background remain the same.

\begin{table}
	\def~{\hphantom{0}}
	\caption{Comparison of average adjusted rand index (ARI) $\times 100$ under heterogeneous backgrounds.
		Numbers within parentheses are empirical standard deviations of ARI $\times 100$. }
	 \begin{center}
		\scalebox{0.7}{
	\begin{tabular}{lccccccc} \\
		\hline \hline
		&\multicolumn{3}{c}{With Logistic Models}&&\multicolumn{3}{c}{Without Logistic Models} \\
		\cline{2-4}  \cline{6-8}
		$p_{11}$  &Poisson & Multinomial & Robust && Poisson & Multinomial & Robust \\
		\hline
		\multicolumn{8}{l}{$62\%$ Background Nodes} \\
		15	&	20	(11)	&	58	(14)	&	54	(21)	&&	15	(6)	&	17	(8)	&	12	(10)	\\
		16	&	23	(13)	&	65	(9)	&	63	(18)	&&	18	(5)	&	23	(10)	&	17	(12)	\\
		17	&	25	(13)	&	70	(8)	&	69	(16)	&&	20	(5)	&	30	(12)	&	24	(17)	\\
		18	&	29	(14)	&	74	(6)	&	76	(11)	&&	23	(5)	&	39	(13)	&	31	(21)	\\
		19	&	35	(18)	&	78	(5)	&	81	(8)	&&	24	(5)	&	50	(12)	&	43	(26)	\\
		20	&	39	(20)	&	80	(5)	&	85	(6)	&&	27	(6)	&	57	(12)	&	50	(27)	\\
		21	&	43	(23)	&	83	(5)	&	88	(5)	&&	29	(5)	&	63	(10)	&	61	(27)	\\
		22	&	48	(25)	&	85	(4)	&	91	(3)	&&	30	(6)	&	66	(10)	&	71	(25)	\\
		23	&	53	(27)	&	86	(4)	&	93	(3)	&&	32	(7)	&	69	(10)	&	78	(22)	\\
		24	&	60	(29)	&	87	(4)	&	94	(2)	&&	34	(9)	&	72	(10)	&	84	(19)	\\
		25	&	67	(30)	&	89	(4)	&	95	(2)	&&	37	(13)	&	74	(10)	&	89	(15)	\\
		\hline
		\multicolumn{8}{l}{$50\%$ Background Nodes} \\
		15	&	62	(19)	&	73	(5)	&	74	(5)	&&	34	(12)	&	44	(9)	&	42	(12)	\\
		16	&	70	(15)	&	77	(4)	&	79	(4)	&&	41	(15)	&	53	(9)	&	55	(11)	\\
		17	&	75	(14)	&	81	(4)	&	83	(4)	&&	46	(15)	&	62	(8)	&	66	(8)	\\
		18	&	81	(10)	&	84	(4)	&	86	(3)	&&	52	(15)	&	69	(6)	&	74	(7)	\\
		19	&	85	(9)	&	86	(4)	&	89	(3)	&&	58	(15)	&	74	(6)	&	80	(6)	\\
		20	&	89	(7)	&	89	(3)	&	92	(3)	&&	63	(17)	&	78	(5)	&	85	(5)	\\
		21	&	92	(4)	&	90	(3)	&	93	(2)	&&	72	(18)	&	82	(5)	&	89	(3)	\\
		22	&	95	(2)	&	92	(3)	&	95	(2)	&&	82	(16)	&	84	(5)	&	92	(2)	\\
		23	&	96	(2)	&	93	(2)	&	96	(2)	&&	88	(15)	&	87	(4)	&	94	(2)	\\
		24	&	97	(2)	&	94	(2)	&	97	(1)	&&	93	(10)	&	88	(4)	&	96	(2)	\\
		25	&	98	(1)	&	94	(2)	&	98	(1)	&&	96	(7)	&	90	(4)	&	97	(1)	\\
		\hline
		\multicolumn{8}{l}{$38\%$ Background Nodes} \\
		15	&	81	(5)	&	82	(4)	&	82	(4)	&&	65	(8)	&	66	(6)	&	67	(7)	\\
		16	&	85	(4)	&	85	(3)	&	86	(3)	&&	71	(7)	&	72	(5)	&	74	(5)	\\
		17	&	89	(3)	&	88	(3)	&	89	(3)	&&	77	(7)	&	78	(5)	&	81	(4)	\\
		18	&	91	(3)	&	90	(3)	&	91	(2)	&&	83	(6)	&	82	(4)	&	85	(4)	\\
		19	&	93	(2)	&	92	(3)	&	94	(2)	&&	88	(4)	&	86	(4)	&	90	(3)	\\
		20	&	95	(2)	&	93	(2)	&	95	(2)	&&	91	(3)	&	88	(3)	&	92	(3)	\\
		21	&	96	(2)	&	94	(2)	&	97	(2)	&&	94	(2)	&	90	(3)	&	94	(2)	\\
		22	&	98	(1)	&	95	(2)	&	98	(1)	&&	96	(2)	&	92	(3)	&	96	(2)	\\
		23	&	98	(1)	&	96	(2)	&	98	(1)	&&	97	(2)	&	93	(2)	&	97	(1)	\\
		24	&	99	(1)	&	97	(2)	&	99	(1)	&&	98	(1)	&	94	(2)	&	98	(1)	\\
		25	&	99	(1)	&	97	(1)	&	99	(1)	&&	99	(1)	&	95	(2)	&	99	(1)	\\
		\hline \hline
	\end{tabular} }
\end{center}
	\label{table2}
	
\end{table}

\begin{figure}
	\begin{center}
		\includegraphics[width=400pt]{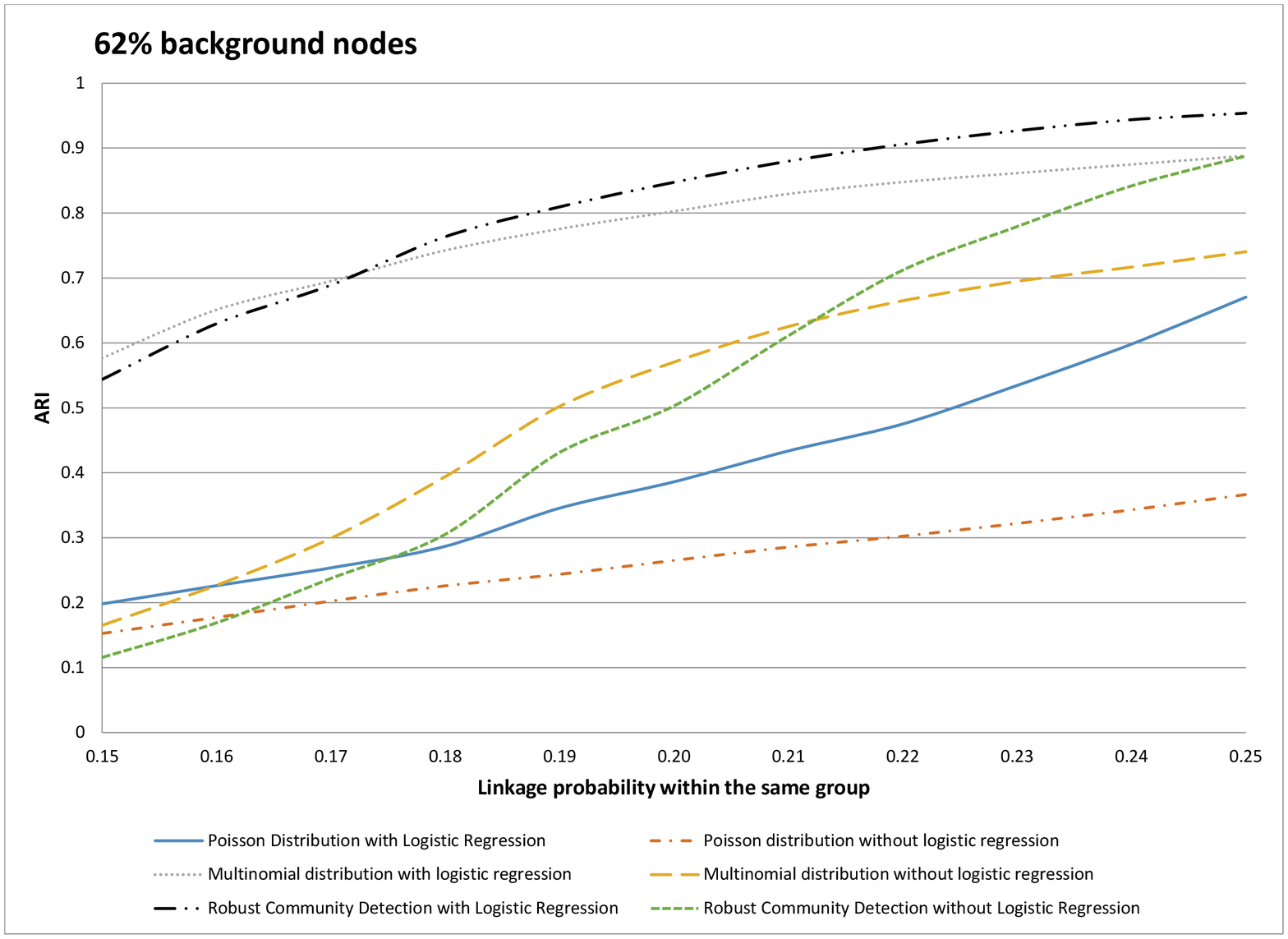}
	\end{center}
	\caption{Comparison of the average ARI for Poisson pseudo-likelihood, multinomial pseudo-likelihood and robust community detection with and without logistic regressions under 62$\%$ of background nodes. This figure appears in color in the electronic version of this article.}
	\label{fig1}
\end{figure}
\begin{figure}
	\begin{center}
		\includegraphics[width=400pt]{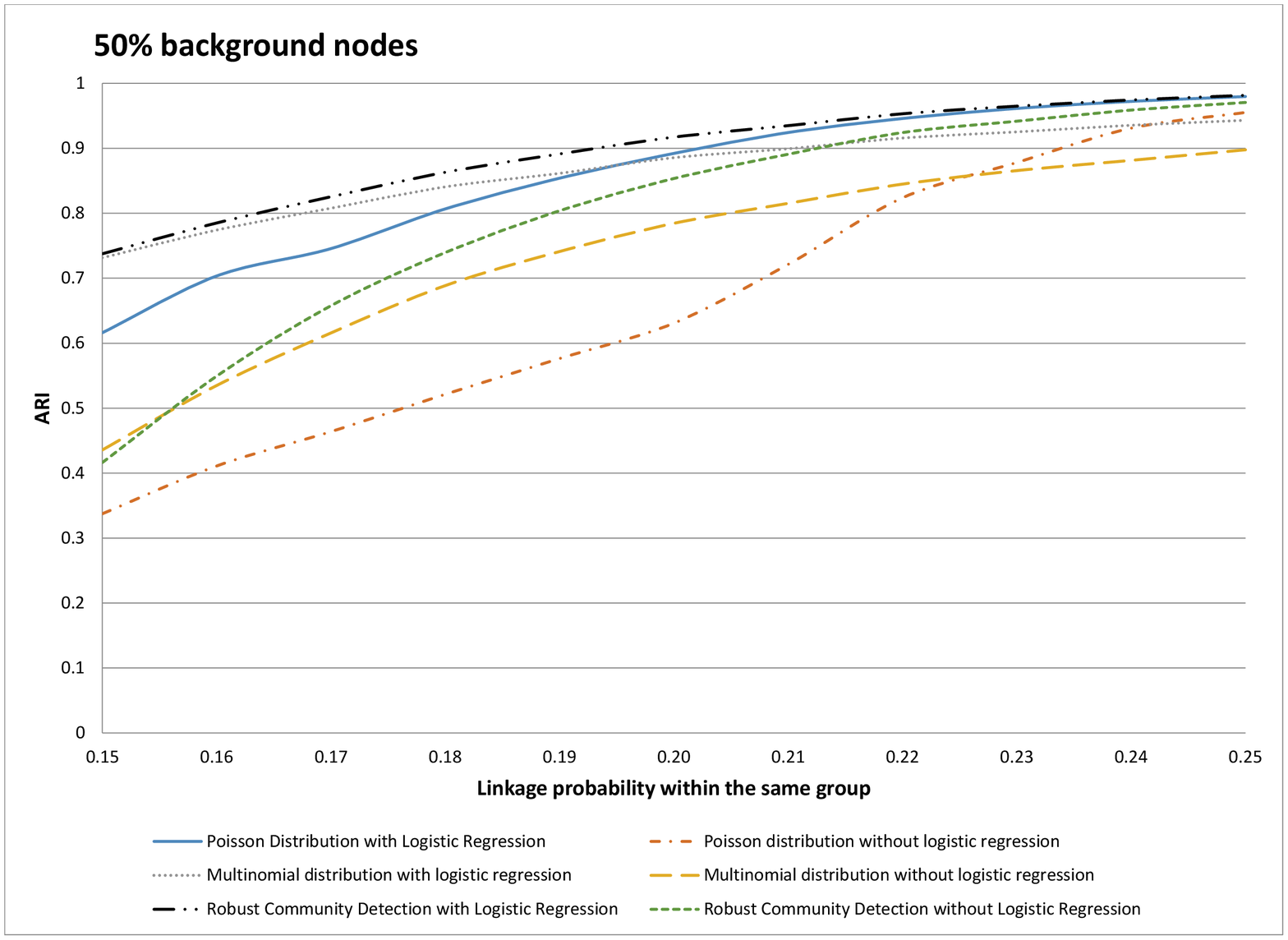}
	\end{center}
	\caption{Comparison of the average ARI for Poisson pseudo-likelihood, multinomial pseudo-likelihood and robust community detection with and without logistic regressions under 50$\%$ of background nodes. This figure appears in color in the electronic version of this article.}
	\label{fig2}
\end{figure}
\begin{figure}
	\begin{center}
		\includegraphics[width=400pt]{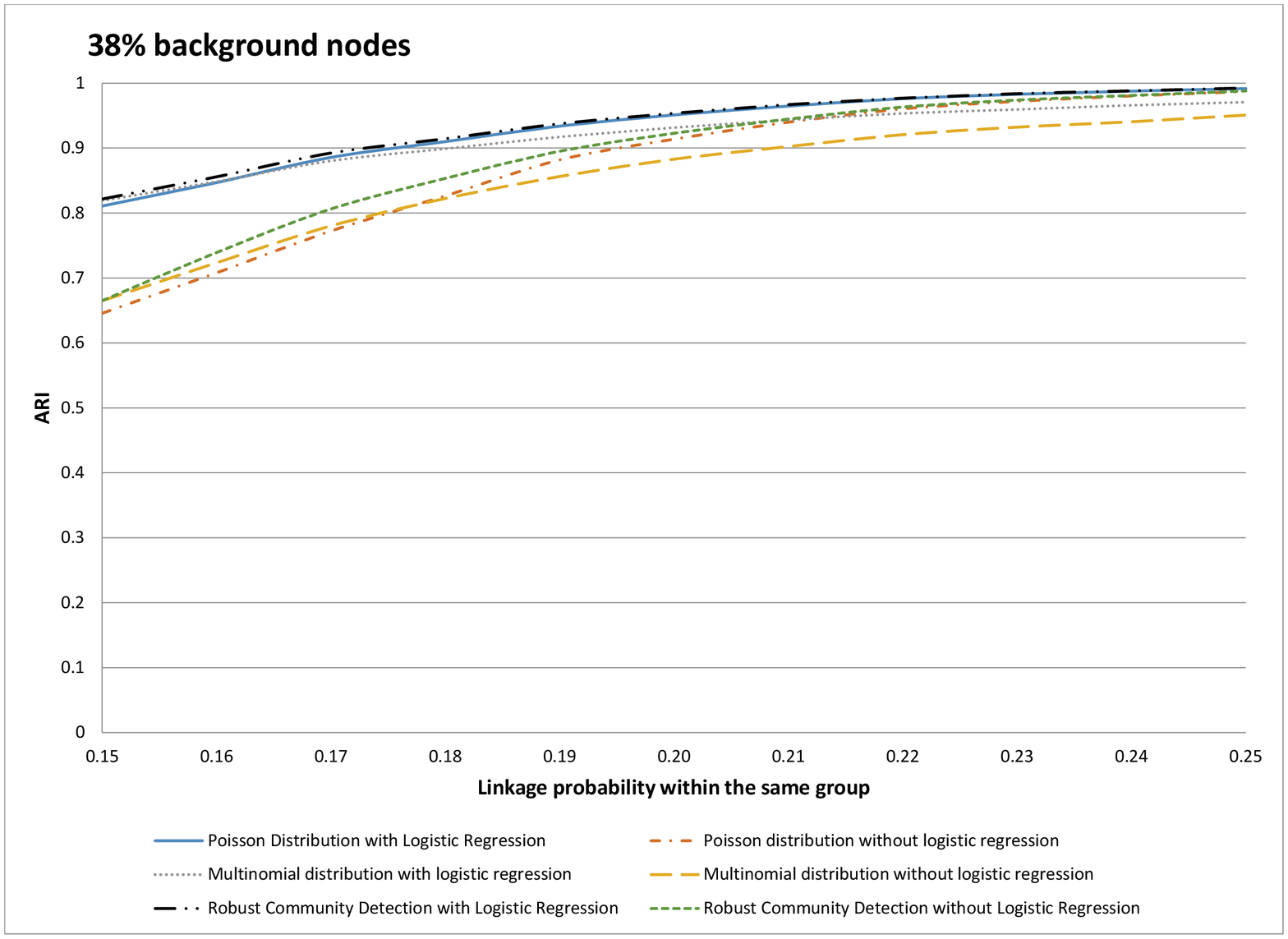}
	\end{center}
	\caption{Comparison of the average ARI for Poisson pseudo-likelihood, multinomial pseudo-likelihood and robust community detection with and without logistic regressions under 38$\%$ of background nodes. This figure appears in color in the electronic version of this article.}
	\label{fig3}
\end{figure}

The \textsc{ari} of the six methods are shown in Table \ref{table2} and Figures \ref{fig1} - \ref{fig3}.
Similar to what we observed in Table \ref{table1}, the average ARIs of all methods increases as the linkage probability within community increases, or as the percentage of background nodes decreases. And the method with logistic regression outperforms their counterparts without logistic regression. The robust method with logistic regression gives the best performance in most scenarios. The Poisson pseudo-likelihood has the worst performance when the stochastic blockmodel assumption is violated in the heterogeneous background. Especially, under the case of high percentage of background nodes, the Poisson pseudo-likelihood performs poorly even when the linkage probability within community is much higher than the linkage probability between communities. The multinomial pseudo-likelihood slightly outperforms the robust method when the percentage of background nodes is high, in which case the robust method discards lots of information, while the multinomial pseudo-likelihood (or correspondingly degree corrected stochastic blockmodel) accounts for high variations on degrees. On the other hand, the robust method outperforms the multinomial pseudo-likelihood in all the other cases. In summary, the robust method has the best performance in terms of both accuracy and efficacy in almost all the setups we examined regardless the data follows stochastic blockmodels or not. In the only exception where the multinomial pseudo-likelihood method with logistic regression performs slightly better, the discrepancies between the two methods are small. Therefore, the robust community detection method is our recommended method.

In real applications, the number of communities is often unknown a priori. \citet{saldana2017many} proposed a modified Bayesian information criterion (BIC) for community detection:

\begin{align}
-2\mathcal{L} (\V{\hat{\beta}},\V{\hat{\pi}},\hat{P} ; \V{\hat{c}},A)+\frac{(K+1)(K+2)}{2} \log \left ( \frac{n(n-1)}{2} \right ). \nonumber
\end{align}
\citet{daudin2008mixture} proposed another model selection criterion -- integrated classification likelihood (ICL) with a heavier penalty:
\begin{align}
-2\mathcal{L} (\V{\hat{\beta}},\V{\hat{\pi}},\hat{P} ; \V{\hat{c}},A)+\frac{(K+1)(K+2)}{2} \log \left ( \frac{n(n-1)}{2} \right ) + K \log(n) . \nonumber
\end{align}

We use simulation studies to verify the performance of BIC and ICL for our case. Since BIC and ICL are designed for the stochastic blockmodel, we compute ${L} (\V{\hat{\beta}},\V{\hat{\pi}},\hat{P} ; \V{\hat{c}},A)$ by \eqref{block} in the present studies although $\V{\hat{c}}$ is estimated by the robust method.

We follow the aforementioned setup for heterogeneous background nodes and only consider the case with 50\% background nodes. For each network, we vary the assumed number of communities from $1$ to $8$, and report in the first three columns of Table \ref{model_selection} the percentages of selected numbers of communities in 50 replicates by BIC and ICL, respectively. Both BIC and ICL perfectly identifies the true community number ($K=2$).

In the last set of simulation studies, we consider the model selection for networks with 5 clusters plus the background. The setup is the same as the previous study except that $n=1000$ and $\beta_0=1$. With this setup, the average size of clusters is approximately 125 as in the previous study. The results are shown in the last three column of Table \ref{model_selection}: BIC and ICL almost perfectly identifies the community number except for one replicate.

\begin{table}
	\def~{\hphantom{0}}
	\caption{Proportions of the Numbers of Communities Selected by BIC and ICL}
	\begin{center}
	\begin{tabular}{ccccccc} \\
		\hline
		\multicolumn{3}{c}{$K=2$} && \multicolumn{3}{c}{$K=5$} \\
		$\hat{K}$ & BIC  & ICL  && $\hat{K}$ & BIC  & ICL \\
		\hline
		1 & 0 & 0 && 1 & 0 & 0 \\
		\textbf{2} & \tb{1} & \tb{1} && 2 & 0 & 0   \\
		3 & 0 & 0 && 3 & 0 & 0  \\
		4 & 0 & 0 && 4 & 0 & 0  \\
		5 & 0 & 0 && \tb{5} & \tb{0.98} & \tb{0.98} \\
		6 & 0  & 0 && 6 & 0.02  & 0.02 \\
		7 & 0  & 0 && 7 & 0  & 0 \\
		8 &  0      & 0 &&  8 &  0      & 0 \\
		\hline
	\end{tabular}
\end{center}
	\label{model_selection}
\end{table}

\section{Application}

With the development of improved sequencing techniques, more and more de novo mutations in candidate genes associated with neurodevelopmental or neuropshychiatric diseases are being reported. Here we focus on autism spectrum disorder and related neurological disorders. Most identified de novo mutations are rare and patients with the same clinical symptoms often carry heterogeneous mutation loci on different genes. Most probably, the pathophysiology mechanism underpinning autism involves perturbed molecular pathways. There is evidence of enrichment of de novo mutations in gene groups connected by protein-protein interactions, co-expression patterns, or pathways defined by common functions, annotations or evolutional patterns \citep{Allen13}. Our study targets at interactive groups of biomarkers that form biological pathways related to autism origination and development. 

Autism and related disorder data from \citet{Hormozdiari15} are employed, which reports four types of information (clinically diagnosed disease status, RNA expression levels, de novo mutations, protein-protein interactions) from three major data consortiums including BrainSpan Atlas, published autism studies, protein-protein interaction databases. There are 52,801 verified protein-protein interaction links and 192,499 mRNA pairs with Pearson's correlation coefficient between their expression levels higher than 0$\cdot$5, with an overlap of 1060 links.  Together, there are 244,240 unique links from both data sources. These links involve 13,243 genes. \citet{Hormozdiari15} further gathered the de novo mutation and length information on 796 out of the 13,243 genes. In total, 796 genes with de novo mutations are employed in our analysis with 1334 mutual links between them, among which 602 genes have at least one link and 194 have none.


Synonymous mutations that differ at the DNA level but produce the same protein products are excluded. The frequencies of each type of mutation in a gene in all cases are summed up as well as the total number in the controls. Two covariates are employed in estimating the probability that a gene is involved in the occurrence or progression of autism and related neurological disorders -- frequency of missense or loss of function mutations in cases, and number of mutations in controls. The choice of the covariates is based on biological beliefs on their involvement on autism development, hence decided a priori.  

As in the simulation study, the BIC \citep{saldana2017many} and ICL \citep{daudin2008mixture} select the same number of communities for this data -- five autism related modules plus one irrelevant background group produces.

We then use three community detection methods to cluster genes into seven clusters: the robust community detection (Section 4), the pseudo-likelihood method with nodal covariates (Section 3) and the standard stochastic blockmodel fitted by the profile-likelihood \citep{bickel2009nonparametric}. We run the algorithm with a number of random initial values for the pseudo-likelihood method with nodal covariates and pick the solution with the largest value of the likelihood \eqref{block}. For a fair comparison, we use this solution as the initial value for the other two methods.

Table 4 shows the estimated link probabilities within gene groups and between gene group pairs for the five autism-related gene groups (group 1-5) and the background gene group (group 6). We compare the estimates from the three methods side by side. According to the table, the standard stochastic blockmodel classifies the nodes with zero connection as a cluster. The pseudo-likelihood method with nodal covariates gives a very similar partition. The adjusted Rand index for the partitions of these two methods are 0.963. Higher values of this index indicate better agreement and 1 means perfect agreement \citep{hubert1985comparing}. The robust method gives a more different partition. The adjusted Rand index for the robust method and the standard stochastic blockmodel is 0.635, which may result from the fact that the robust method allows heterogeneous linkage rates in background genes while the two stochastic blockmodels do not.  Table \ref{table_density} shows the estimated link probabilities for the three methods. Furthermore, the estimated odds ratio from the robust method for mutation numbers in cases and mutation numbers in controls are 1.4874 (P-value=0.2808) and  -1.0335  (P-value=0.0173), respectively. On the contrary, the two odds ratio estimates are  -0.5332 (P-value=0.0419) and  -0.2464  (P-value=0.5770) from the stochastic blockmodels, which disagrees with the prior that genes with higher number of mutations in cases are more likely to be related to neurological disorders. Therefore, we employ the results from the robust method.

\begin{table}
	\def~{\hphantom{0}}
	\caption{Estimated Link Probabilities between Groups}
	\begin{center}
	\begin{tabular}{cccccc}\\
		\hline
		\multicolumn{5}{l}{Stochastic Blockmodel:} \\
		\multicolumn{5}{c}{Group 1--5}  &\textit{Group 6}\\
		\hline
		0.377 & 0.014  &  0 & 0.002 & 0.003 & 0.002 \\
		0.014 & 0.410  &  0 & 0.001 & 0.034 & 0.069 \\
		0.000 & 0.000  &  0 & 0.000 & 0.000 & 0.000 \\
		0.002 & 0.001  &  0 & 0.009 & 0.004 & 0.000 \\
		0.003 & 0.034  &  0 & 0.004 & 0.027 & 0.081 \\
		0.002 & 0.069  &  0 & 0.000 & 0.081 & 0.529 \\
		\hline
		\multicolumn{5}{l}{Pseudo Likelihood with Covariates:} \\
		\multicolumn{5}{c}{Group 1--5}  &\textit{Group 6}\\
		\hline
		0.414 & 0.014 &   0 & 0.002 & 0.004 & 0.002 \\
		0.014 & 0.433  &   0 & 0.003 & 0.046 & 0.065 \\
		0.000 & 0.000 &    0 & 0.000 & 0.000 & 0.000 \\
		0.002 & 0.003  &   0 & 0.007 & 0.006 & 0.003 \\
		0.004 & 0.046  &   0 & 0.006 & 0.035 & 0.095 \\
		0.002 & 0.065   &  0 & 0.003 & 0.095 & 0.514 \\
		\hline
		\multicolumn{5}{l}{Robust Community Detection:} \\
		\multicolumn{5}{c}{Group 1--5}  &\textit{Group 6}\\
		\hline
		0.414 & 0.002 & 0.000 & 0.004 & 0.012 & 0.000 \\
		0.002 & 0.618 & 0.000 & 0.006 & 0.190 & 0.028 \\
		0.000 & 0.000 & 0.000 & 0.001 & 0.000 & 0.003 \\
		0.004 & 0.006 & 0.001 & 0.009 & 0.009 & 0.027 \\
		0.012 & 0.190 & 0.000 & 0.009 & 0.087 & 0.055 \\
		0.000 & 0.028 & 0.003 & 0.027 & 0.055 & 0.104 \\
		\hline
	\end{tabular}
\end{center}
	\label{table_density}
\end{table}

The gene set enrichment analysis (GSEA) of the selected gene modules compared with the curated gene sets in the Molecular Signatures Database are listed in Table 5. P-values are calculated assuming a hypergeometric distribution for the number of overlapping genes between the selected group and the curated gene set. Given the large number of multiple comparisons, stringent P-value threshold $10^{-10}$ is employed. The five autism-related groups overlaps significantly with gene sets in essential cellular functions or abnormal conditions such as cancer, apoptosis, cell structure, circulatory system, nervous system, multicellular organismal development. Group four overlaps with gene sets related to neurological functions or disorders. Gene set ``GO NEUROGENESIS" are genes involved in generation of cells within the nervous system. Gene set ``GO REGULATION OF NERVOUS SYSTEM DEVELOPMENT" concerns processes that modulate the frequency, rate or extent of nervous system development, the origin and formation of nervous tissue. Gene sets ``GO NEURON PROJECTION" and ``GO SYNAPSE" are composed of genes involved in nerve cell prolongation and nerve fiber junction, respectively. Furthermore, our results are compared to those from the Merging Aaffected Genes into Integrated-Nnetworks method in \citet{Hormozdiari15}. The Merging Affected Genes into Integrated-Networks method was not able to detect group four. P-values from gene set enrichment analysis for the two best sets identified by their method against known neurodevelopmental diseases sets are 4$\cdot$2$\times 10^{-5}$ and 1$\cdot$0$\times 10^{-4}$, failing to reach the $10^{-10}$ threshold.

\begin{table}
	\def~{\hphantom{0}}
	\caption{Gene Set Enrichment Analysis of Selected Groups. The first column is the group number identified by the proposed method; Size refers to the number of genes in the identified group, or gene set in the GSEA or their overlap.}%
		\scalebox{0.6}{
		\begin{tabular}{llccccc}
			\\
			\hline
			Group &Gene Set    &Group     &GeneSet  & Overlap &Nominal  & FDR \\
			Number &Name         &Size    & Size       & Size & P-value  & q-value  \\
			\hline
			1&	FISCHER DREAM TARGETS                               &   18  &929	&16	     &1$\cdot$01 $\times 10^{-25}$	&1$\cdot$07 $\times 10^{-21}$\\
			1&	GOBERT OLIGODENDROCYTE DIFFERENTIATION              &	18	&570    &11	     &2$\cdot$86 $\times 10^{-17}$	&1$\cdot$52 $\times 10^{-13}$\\
			1&	DUTERTRE ESTRADIOL RESPONSE 24HR UP                 &	18	&324	&9	     &1$\cdot$77 $\times 10^{-15}$	&6$\cdot$29 $\times 10^{-12}$\\
			1&	FISCHER G2 M CELL CYCLE                             &	18	&225	&8	     &1$\cdot$22 $\times 10^{-14}$	&3$\cdot$26 $\times 10^{-11}$\\
			1&	PUJANA BRCA2 PCC NETWORK                            &   18	&423    &9	     &1$\cdot$97 $\times 10^{-14}$	&4$\cdot$19 $\times 10^{-11}$\\
			1&	PUJANA XPRSS INT NETWORK                            &   18	&168	&7	     &2$\cdot$37 $\times 10^{-13}$	&4$\cdot$20 $\times 10^{-10}$\\
			1&	GEORGES TARGETS OF MIR192 AND MIR215                &	18	&893	&10	     &2$\cdot$78 $\times 10^{-13}$	&4$\cdot$23 $\times 10^{-10}$\\
			1&	NUYTTEN EZH2 TARGETS DN                             &	18	&1024	&10    	 &1$\cdot$08 $\times 10^{-12}$	&1$\cdot$43 $\times 10^{-9}$\\
			1&	PUJANA CHEK2 PCC NETWORK                            &	18	&779	&9	     &4$\cdot$68 $\times 10^{-12}$	&5$\cdot$54 $\times 10^{-9}$\\
			1&	GO CHROMOSOME                                       &   18  &880	&9       &1$\cdot$38 $\times 10^{-11}$	&1$\cdot$48 $\times 10^{-8}$\\
			
			2&	GO CHROMOSOME ORGANIZATION                          &   29	&1009	&11    	 &1$\cdot$31 $\times 10^{-11}$	&8$\cdot$62 $\times 10^{-8}$\\
			2&	GRAESSMANN APOPTOSIS BY DOXORUBICIN DN              &	29	&1781	&13 	 &1$\cdot$62 $\times 10^{-11}$	&8$\cdot$62 $\times 10^{-8}$\\
			2&	DACOSTA UV RESPONSE VIA ERCC3 DN                    &   29  &855	&10	     &6$\cdot$86 $\times 10^{-11}$	&2$\cdot$44 $\times 10^{-7}$\\
			
			3&	GO INTRINSIC COMPONENT OF PLASMA MEMBRANE           &   518	&1649	&71	     &4$\cdot$89 $\times 10^{-22}$	&5$\cdot$21 $\times 10^{-18}$\\
			3&	GO RIBONUCLEOTIDE BINDING                           &	518	&1860	&75	     &1$\cdot$37 $\times 10^{-21}$	&7$\cdot$29 $\times 10^{-18}$\\
			3&	GO TRANSPORTER ACTIVITY                             &   518 &1276	&60	     &1$\cdot$53 $\times 10^{-20}$	&5$\cdot$43 $\times 10^{-17}$\\
			3&	DODD NASOPHARYNGEAL CARCINOMA UP                    &	518	&1821   &71	     &1$\cdot$16 $\times 10^{-19}$	&3$\cdot$10 $\times 10^{-16}$\\
			3&	GO ION TRANSPORT                                    &   518	&1262	&58	    &2$\cdot$04 $\times 10^{-19}$	&4$\cdot$34 $\times 10^{-16}$\\
			3&	GO CELL PROJECTION                                  &   518	&1786	&69	    &6$\cdot$53 $\times 10^{-19}$	&1$\cdot$07 $\times 10^{-15}$\\
			3&	GO ADENYL NUCLEOTIDE BINDING                        &   518	&1514	&63	    &7$\cdot$01 $\times 10^{-19}$	&1$\cdot$07 $\times 10^{-15}$\\
			3&	GO TRANSMEMBRANE TRANSPORTER ACTIVITY               &   518	&997	&49	    &1$\cdot$07 $\times 10^{-17}$	&1$\cdot$43 $\times 10^{-14}$\\
			3&	GO PLASMA MEMBRANE REGION                           &   518	&929	&47	    &1$\cdot$68 $\times 10^{-17}$	&1$\cdot$99 $\times 10^{-14}$\\
			3&	GO TRANSMEMBRANE TRANSPORT                          &   518	&1098	&51	    &2$\cdot$29 $\times 10^{-17}$	&2$\cdot$44 $\times 10^{-14}$\\
			
			4&	GO CELL PROJECTION                                  &   138	&1786	&39	    &5$\cdot$66 $\times 10^{-23}$	&6$\cdot$03 $\times 10^{-19}$\\
			4&	GO REGULATION OF CELL DEVELOPMENT                   &   138	&836	&28	    &2$\cdot$62 $\times 10^{-21}$	&1$\cdot$39 $\times 10^{-17}$\\
			4&	GO CIRCULATORY SYSTEM DEVELOPMENT                   &   138	&788	&27	    &8$\cdot$15 $\times 10^{-21}$	&2$\cdot$89 $\times 10^{-17}$\\
			4&	GO NEUROGENESIS                                     &   138	&1402	&32	    &2$\cdot$29 $\times 10^{-19}$	&6$\cdot$11 $\times 10^{-16}$\\
			4&	GO REGULATION OF MULTICELLULAR ORGANISMAL DEVELOPMENT&  138	&1672	&34	    &4$\cdot$94 $\times 10^{-19}$	&1$\cdot$05 $\times 10^{-15}$\\
			4&	GO VASCULATURE DEVELOPMENT                          &   138	&469	&21	    &1$\cdot$12 $\times 10^{-18}$	&1$\cdot$98 $\times 10^{-15}$\\
			4&	GO REGULATION OF CELL DIFFERENTIATION               &   138	&1492	&32	    &1$\cdot$40 $\times 10^{-18}$	&2$\cdot$12 $\times 10^{-15}$\\
			4&	GO REGULATION OF NERVOUS SYSTEM DEVELOPMENT         &   138	&750	&24	    &6$\cdot$78 $\times 10^{-18}$	&9$\cdot$03 $\times 10^{-15}$\\
			4&	GO NEURON PROJECTION                                &   138	&942	&26	    &8$\cdot$96 $\times 10^{-18}$	&1$\cdot$06 $\times 10^{-14}$\\
			4&	GO SYNAPSE                                          &   138	&754	&23	    &9$\cdot$88 $\times 10^{-17}$	&1$\cdot$04 $\times 10^{-13}$\\
			
			5&	GO CHROMOSOME ORGANIZATION                          &   62	&1009	&20	    &2$\cdot$17 $\times 10^{-18}$	&2$\cdot$31 $\times 10^{-14}$\\
			5&	GO CHROMATIN MODIFICATION                           &   62	&539	&15	    &5$\cdot$06 $\times 10^{-16}$	&2$\cdot$69 $\times 10^{-12}$\\
			5&	GO CHROMATIN ORGANIZATION                           &   62	&663	&15	    &1$\cdot$04 $\times 10^{-14}$	&3$\cdot$69 $\times 10^{-11}$\\
			5&	PILON KLF1 TARGETS DN                               &   62	&1972	&19	    &7$\cdot$01 $\times 10^{-12}$	&1$\cdot$87 $\times 10^{-8}$\\
			5&	GO CELL CYCLE                                       &   62	&1316	&16	    &1$\cdot$48 $\times 10^{-11}$	&3$\cdot$15 $\times 10^{-8}$\\
			5&	GO COVALENT CHROMATIN MODIFICATION                  &   62	&345	&10	    &3$\cdot$80 $\times 10^{-11}$	&6$\cdot$74 $\times 10^{-8}$\\
			5&	DACOSTA UV RESPONSE VIA ERCC3 DN                    &   62	&855	&13	    &1$\cdot$05 $\times 10^{-10}$	&1$\cdot$50 $\times 10^{-7}$\\
			5&	GO NUCLEAR CHROMOSOME                               &   62	&523	&11	    &1$\cdot$13 $\times 10^{-10}$	&1$\cdot$50 $\times 10^{-7}$\\
			5&	GO CELL CYCLE PROCESS                               &   62	&1081	&14	    &1$\cdot$49 $\times 10^{-10}$	&1$\cdot$59 $\times 10^{-7}$\\
			5&	GO CHROMOSOME                                       &   62	&880	&13	    &1$\cdot$49 $\times 10^{-10}$	&1$\cdot$59 $\times 10^{-7}$\\
			\hline
			\vspace{10mm}
			\footnote{}
			
		\end{tabular}
	   
	}
		\label{table5}
		\vspace{1ex}

\end{table}

\section{Discussion}

A major improvement of the proposed method over previous ones is the integration of network topology and auxiliary node information. The proposed analysis pools rich epigenomic information from heterogeneous online resources, such as expression/co-expression profiles from BrainSpan Atlas, de novo mutations in cases and controls from autism or related neurological disorder studies, protein-protein interactions in protein databases. Although these three types of information are measured on different cohorts, they describe distinct aspects of the candidate genes. They can be linked by unique genes, which are the unit of our analysis. In the era of big data, statistical methods need not be restricted to one data source or single clinical trial. Instead, methods should incorporate information from multiple related resources.

The estimation method is non-standard. For a fixed initial label assignment, we use the expectation-maximization algorithm to fit a pseudo-likelihood. Then the label assignment is updated according to the expectation-maximization results, and used as initial label assignment in the next iteration. Taking advantage of the pseudo-likelihood, we are able to allow heterogeneous linkage probabilities in the background. The consistency of the label assignments is proved for a simple version of this complicated procedure -- one-step expectation-maximization. Further research is needed to understand the statistical properties of the algorithm in more complex settings. Furthermore, the pseudo-likelihood approach for the robust setup is very unique. Usually, a regularization term is used to penalize the log-likelihood when the MLE is degenerate, for instance, the roughness penalty in smoothing splines and $L_1$ penalty in lasso for ``large-$p$-small-$n$'' problems. Our approach for the robust setup does not follow exactly the ``loss+penalty'' framework. Due to the special two-dimensional structure (adjacency matrix) of network data, the pseudo-likelihood with the noisy data being removed can give accurate community labels while the MLE of the true likelihood is degenerate. In network analysis the pseudo-likelihood approach not only provides computational efficiency but can also serve as a convenient likelihood formulation that can discard some columns of variables (noise/background) while still remain a valid likelihood because all rows of observations (nodes/genes) are kept. The full likelihood will fail in this case.

Researchers have suggested that a node may belong to multiple communities in a biological networks. For example, \cite{airoldi2008mixed} proposed a mixed membership stochastic blockmodels and applied this model into a network of protein-protein interactions. We will explore the extension of the logistic regression augmented model to overlapping community detection in our future work.

\appendix
\section{Proof of Identifiability of Pseudo-likelihoods}
Both the non-robust and robust version of the pseudo-likelihoods are a combination of a logistic regression and a mixture/compound multivariate Poisson model. Specifically, the robust pseudo-log-likelihood is the sum of the log-likelihood from the logistic regression and the log-likelihood from the mixture of $K$-variate (we only used $K$ columns) while the non-robust version includes $K+1$ variates. The identifiability of mixture Poisson models is a very classical result, for instance see \cite{feller2015general}. The identifiability of our pseudo-likelihood can be derived by the identifiability of mixture Poisson when $n\geq P$. For notational convenience, we only consider the case that the covariates $x_i$ is univariate and the multivariate case can be proved similarly. The joint distribution of the first two columns in the robust likelihood is 
\begin{align}\label{Poisson}
& \prod_{i=1}^2  \left \{  \sum_{l=1}^K \frac{e^{\beta_0+x_i \beta_1 }}{1+e^{\beta_0+x_i \beta_1}}\pi_l e^{-\mu_l} \left ( \prod_{k=1}^{K}  \lambda_{lk}^{b_{ik}} \right ) +\frac{1}{1+e^{\beta_0+x_i \beta_1 }} e^{-\mu_{K+1}} \left ( \prod_{k=1}^{K}  \lambda_{K+1,k}^{b_{ik}} \right ) \right \} 
\end{align}
For $i=1,2$, let $\rho_{il}=\frac{e^{\beta_0+x_i \beta_1 }}{1+e^{\beta_0+x_i \beta_1 }}\pi_l, \,\,l=1,...,K$ and $\rho_{i,K+1}=\frac{1}{1+e^{ \beta_0+x_i \beta_1 }}$.

Then \eqref{Poisson} becomes 
\begin{align*}
& \prod_{i=1}^2  \left \{  \sum_{l=1}^{K+1} \rho_{il} e^{-\mu_l} \left ( \prod_{k=1}^{K}  \lambda_{lk}^{b_{ik}} \right ) \right \}.
\end{align*}
According to the identifiability of mixture Poisson, $\rho_{il}$ is uniquely determined for $i=1,2$, $l=1,...,K+1$. Finally,
\begin{align*}
\rho_{1,K+1} &= \frac{1}{1+e^{ \beta_0+x_1 \beta_1 }} \\
\rho_{2,K+1} &= \frac{1}{1+e^{ \beta_0+x_2 \beta_1 }} 
\end{align*}
can uniquely determine $\beta_0,\beta_1$ and hence $\pi_1,...,\pi_K$ are also unique.
\section{Proof of Lemma 5.1}
Recall that $\hat{\beta}_0$ and $\hat{\beta}_1$ can be obtained by
\begin{align}
(\hat{\beta}_0, \hat{\beta}_1) & = \argmax_{\beta_0,\beta_1} \sum_{i=1}^n \left \{   y_i(\beta_0+x_i\beta_1)- \log(1+e^{\beta_0+x_i\beta_1} )  \right \}. \nonumber
\end{align}
Taking derivative of the log-likelihood above with respect to $\beta_0$ and $\beta_1$, we obtain

\begin{align}
\sum_{i=1}^n \frac{e^{x_i\hat{\beta}_1+\hat{\beta}_0}}{1+e^{x_i\hat{\beta}_1+\hat{\beta}_0}}  & =\sum_{i=1}^n  y_i ,\label{try2} \\
\sum_{i=1}^n \frac{x_i e^{x_i\hat{\beta}_1+\hat{\beta}_0}}{1+e^{x_i\hat{\beta}_1+\hat{\beta}_0}} & = \sum_{i=1}^n  y_i x_i .\label{try3}
\end{align}

Denote $s=(1/n) \sum_{i=1}^n  y_i$. Then $s$ is constant for $\V{e} \in \mathcal{E}$, since $s=n_1 \gamma_1+ n_2(1-\gamma_2)$.

By the defintion of Riemann integral, for sufficiently large $n$,
\begin{align}
(1-\epsilon)s \leq \int_{0}^1 \frac{e^{x\hat{\beta}_1+\hat{\beta}_0}}{1+e^{x\hat{\beta}_1+\hat{\beta}_0}} dx \leq (1+\epsilon)s.  \label{ineq1}
\end{align}

Without loss of generality, we assume $\hat{\beta}_1 \neq 0$, since it is easy to show that $\hat{\beta}_0$ is bounded from \eqref{try3} otherwise.

Under this assumption, the integral in \eqref{ineq1} has a closed form:
\begin{equation}
\int_{0}^1 \frac{e^{x \hat{\beta}_1+\hat{\beta}_0}}{1+e^{x\hat{\beta}_1+\hat{\beta}_0}} dx =\frac{1}{\hat{\beta}_1}\{\log(1+e^{\hat{\beta}_0+\hat{\beta}_1})-\log(1+e^{\hat{\beta}_0})\}.\nonumber
\end{equation}

First we consider the case that $\hat{\beta}_1>0$. According to \eqref{ineq1},
\begin{equation}\label{du}
\log \frac{e^{s(1-\epsilon)\hat{\beta}_1}-1}{e^{\hat{\beta}_1}-e^{s(1-\epsilon) \hat{\beta}_1}} \leq \hat{\beta}_0 \leq \log \frac{e^{s(1+\epsilon)\hat{\beta}_1}-1}{e^{\hat{\beta}_1}-e^{s(1+\epsilon) \hat{\beta}_1}}.
\end{equation}

By \eqref{du}, it is easy to check that
\begin{align}
\lim_{\hat{\beta}_1 \to +\infty} e^{x \hat{\beta}_1+\hat{\beta}_0} & \geq \lim_{\hat{\beta}_1 \to +\infty}\frac{e^{(x+s(1-\epsilon)) \hat{\beta}_1}-e^{x\hat{\beta}_1}}{e^{\hat{\beta}_1}-e^{s(1-\epsilon)\hat{\beta}_1}} = \left \{ \begin{tabular}{cc}
+$\infty$ & if $x>1-s(1-\epsilon)$, \\
0 & if $x<1-s(1-\epsilon)$.
\end{tabular} \nonumber
\right .
\end{align}
Therefore, for sufficiently large $n$,
\begin{equation}\label{du1}
\lim \limits _{\hat{\beta}_1 \to +\infty} \frac{1}{n} \sum_{i=1}^n \frac{x_i e^{x_i\hat{\beta}_1+\hat{\beta}_0}}{1+e^{x_i\hat{\beta}_1+\hat{\beta}_0}}  \geq \lim \limits _{\hat{\beta}_1 \to +\infty} (1-\epsilon) \int_{0}^1 \frac{xe^{x\hat{\beta}_1+\hat{\beta}_0}}{1+e^{x\hat{\beta}_1+\hat{\beta}_0}} dx  \geq (1-\epsilon) \int_{1-s(1-\epsilon)}^1 xdx.
\end{equation}
However,
\begin{equation}\label{du3}
\max \limits _{\V{e} \in \mathcal{F}\cap \mathcal{E}} \frac{1}{n}\sum \limits_{i=1}^n y_i x_i \leq  \frac{1}{n}\sum \limits_{i=n-\hat{n}_1 \tilde{\gamma}_1+1}^n x_i + \frac{1}{n}\sum \limits_{i=\hat{n}_2-\hat{n}_1(1-\tilde{\gamma}_1)+1}^{\hat{n}_2} x_i,
\end{equation}
the right hand side of \eqref{du3} converges to
\begin{align}\label{du4}
\int_{1-\tilde{\gamma}_1 s}^1 xdx+ \int_{1-s-s(1-\tilde{\gamma}_1)}^{1-s} xdx,
\end{align}
and thus it is strictly less than \eqref{du1}. Therefore, there exists $M_1$ such that $\hat{\beta}_1<M_1$  for sufficiently large $n$. Note that $M_1$ only depends on \eqref{du4}, and hence is independent with $n$.

Similarly, when $\hat{\beta}_1 <0$,

\begin{align}
\log \frac{e^{s(1+\epsilon)\hat{\beta}_1}-1}{e^{\hat{\beta}_1}-e^{s(1+\epsilon) \hat{\beta}_1}} \leq \hat{\beta}_0 \leq \log \frac{e^{s(1-\epsilon)\hat{\beta}_1}-1}{e^{\hat{\beta}_1}-e^{s(1-\epsilon) \hat{\beta}_1}}. \label{duNeg}
\end{align}
For sufficiently large $n$,
\begin{align}
\lim \limits _{\hat{\beta}_1 \to -\infty} \frac{1}{n} \sum_{i=1}^n \frac{x_i e^{x_i\hat{\beta}_1+\hat{\beta}_0}}{1+e^{x_i\hat{\beta}_1+\hat{\beta}_0}}  \leq \lim \limits _{\hat{\beta}_1 \to -\infty} (1+\epsilon) \int_{0}^1 \frac{xe^{x\hat{\beta}_1+\hat{\beta}_0}}{1+e^{x\hat{\beta}_1+\hat{\beta}_0}} dx \leq (1+\epsilon) \int_0^{s} xdx.\nonumber
\end{align}

But
\begin{align}
\min \limits _{\V{e} \in \mathcal{F}\cap \mathcal{E}} \frac{1}{n} \sum  \limits_{i=1}^n y_i x_i\geq \frac{1}{n} \sum_{i=1}^{\tilde{\gamma}_2 \hat{n}_1} x_i + \frac{1}{n} \sum_{i=\hat{n}_1+1}^{\hat{n}_1+(1-\tilde{\gamma}_2) \hat{n}_1} x_i \rightarrow \int_0^{ \tilde{\gamma}_2 s } x dx + \int_s^{s+(1-\tilde{\gamma}_2) s} x dx.\nonumber
\end{align}
which implies $\hat{\beta}_1>-M_2$ for a fixed positive value of $M_2$. It implies the solution $(\hat{\beta}_0,\hat{\beta}_1)$ for \eqref{try2} and \eqref{try3} is bounded together with \eqref{du} and \eqref{duNeg}.

%
%
%
%
%

\section{Proof of Theorem 5.1}
With Lemma 5.1, the proof of Theorem 5.1 is closely followed the proof of Proposition 1 and Theorem 3 in \citep{AAA}. We give the details for completeness. We begin with notation. Recall that confusion matrix $R$ defined as $R_{ka}=(1/n)\sum\limits_{i=1}^n 1 (e_i=k,c_i=a)$ is constant in $\mathcal{E}$ and is given by

\begin{center}
	$R=\left(
	\begin{array}{cc}
	\gamma_1\pi_1 & (1-\gamma_2)\pi_2  \\
	(1-\gamma_1)\pi_1 & \gamma_2\pi_2 \\
	\end{array}
	\right)$.\nonumber
\end{center}
Let $\tau=\pi_2/\pi_1$ and define

\begin{align}
u(x)=\frac{(1-\gamma_1)x+\gamma_2\tau}{\gamma_1x+(1-\gamma_2)\tau}, \quad v(x)=u(\frac{1}{x}), \nonumber
\end{align}
and
\begin{align}
F_1(x,y)=\log\frac{1+u(x)}{1+v(y)}, \quad F_2(x,y)=\log\frac{1+[u(x)]^{-1}}{1+[v(y)]^{-1}}. \nonumber
\end{align}

Define the Kullback¨C-Leibler divergence of two Bernoulli distribution with success rates $p$ and $q$ respectively as 
\begin{align}
D(p||q)=p \log\frac{p}{q}+(1-p)\log \frac{1-p}{1-q}. \nonumber
\end{align}
In addition to the conditions listed in the main text of Theorem 2, we need the following regularity condition: There exists $\delta \in (0,1)$ such that
\begin{align}
\frac{\tau}{\rho_1}(1+\delta) \leq \frac{D(\gamma_1||(1-\gamma_2))) }{D((1-\gamma_2)||\gamma_1)} \leq (1-\delta) \rho_2 \tau.\nonumber
\end{align}

Let $\mathcal{C}_{\ell}$ be the set of nodes in true community $\ell$, and $\mathcal{S}_k$ be the set of nodes in community $k$ according to initial labeling $e$. We set $n_{\ell}=|\mathcal{C}_{\ell}|, \hat{n}_k=|\mathcal{S}_k|,$ and $\mathcal{S}_{k\ell}=\mathcal{S}_k \bigcap \mathcal{C}_{\ell}$.

%

Now we consider $i \in \mathcal{C}_1$. Then $\hat{c}_i(\V{e})=1$ if
\begin{equation}
b_{i1}\log\frac{\hat{\theta}_{21}}{\hat{\theta}_{11}}+b_{i2}\log\frac{\hat{\theta}_{22}}{\hat{\theta}_{12}}<\log\frac{\hat{\pi}_{i1}}{1-\hat{\pi}_{i1}}. \nonumber
\end{equation}
Let $\hat{\pi}_{(1)}$ be the smallest value of $\hat{\pi}_{i1} \,\, (i=1,...,n)$.
Define
\begin{align}
\alpha_1 & =\log\frac{\hat{\theta}_{21}}{\hat{\theta}_{11}}, \quad \alpha_2 =\log\frac{\hat{\theta}_{22}}{\hat{\theta}_{12}}, \nonumber
\\
\sigma_j(\V{e}) & =\alpha_11\{e_j=1\}+\alpha_21\{e_j=2\}, \quad (j=1,...,n) \nonumber\\
\hat{\tau}_{(1)} & =  \frac{1-\hat{\pi}_{(1)}}{\hat{\pi}_{(1)}} . \nonumber
\end{align}
So that $\alpha_1b_{i1}+\alpha_2b_{i2}=\sum_j A_{ij}\sigma_j(\V{e})=\xi_i\{\sigma(\V{e})\}$.
Thus, the mis-match ratio over class 1 (with identity permutation) is,
\begin{align}
{M}_{n,1}(\V{e}) & =(1/{n_1})\sum  \limits_{i \in \mathcal{C}_1} 1\{\hat{c}_i(\V{e})\neq 1\} \nonumber\\
& \leq (1/{n_1})\sum \limits_{i \in \mathcal{C}_1} 1\{ \alpha_1 b_{i1}+ \alpha_2 b_{i2} \geq \log\frac{\hat{\pi}_{i1}}{1-\hat{\pi}_{i1}} \} \nonumber\\
& \leq (1/{n_1}) \sum \limits_{i \in \mathcal{C}_1} 1\{ \alpha_1 b_{i1}+ \alpha_2 b_{i2} \geq - \log \hat{\tau}_{(1)} \}\nonumber
\end{align}

By Bernstein inequality, we have

\begin{align}
\textnormal{pr}[\xi_i(\sigma)\ge E\{\xi_i(\sigma)\}+t] \le \exp \left\{  -\frac{t^2/2}{\sum_j \textnormal{var}(A_{ij}\sigma_j)+\norm{\alpha}_{\infty} t/3 } \right\}, \nonumber
\end{align}
where $\norm{\alpha}_{\infty}:=\max{|\alpha_1|, |\alpha_2|}$ and we have used that $|\widetilde{A}_{ij}\sigma_j| \le \norm{\alpha}_{\infty} $ since $i \in \mathcal{C}_1$, then we have
\begin{align}
E[\xi_i(\sigma)] & =\sum \limits _j \sigma_j E[A_{ij}]= \sum \limits_{k=1}^2\sum \limits_{\ell=1}^2\sum \limits_{j}\sigma_j E[A_{ij}]1\{j \in \mathcal{S}_{k\ell}\} \nonumber\\
& = \sum \limits_{k=1}^2\sum \limits_{\ell=1}^2\sum \limits_{j}\alpha_k P_{1\ell}1\{j \in \mathcal{S}_{k\ell}\} =n[\alpha^TR P ]_1 = [\Lambda \alpha]_1. \nonumber
\end{align}

In which  $[\Lambda\alpha]_1$ denotes the value for the first row of $\Lambda\alpha$, so is $[\alpha^TR P ]_1$. By a similar argument,

\begin{align}
\sum \limits_j \mbox{var}(A_{ij}\sigma_j) & =\sum \limits_j \sigma_j^2 \mbox{var}[A_{ij}] \nonumber\\
& \le \sum \limits_j \sigma_j^2 E[A_{ij}] \le \norm{\alpha}_{\infty}\sum \limits_j |\sigma_j|E[A_{ij}]=\norm{\alpha}_{\infty}[\Lambda|\alpha|]_1, \nonumber
\end{align}
where $|\alpha|=(|\alpha_1|,|\alpha_2|)$. Combining what we've got above, we have
\begin{align}
\textnormal{pr}[\xi_i(\sigma)\ge E\{\xi_i(\sigma)\}+t] \le \exp \left[  -\frac{t^2}{2\norm{\alpha}_{\infty} \{  [\Lambda|\alpha|]_1+t/3\} } \right]. \nonumber
\end{align}

Take $t=z_{1,n}=-[\Lambda\alpha]_1-\log \hat{\tau}_{(1)}$. We now show that $-[\Lambda\alpha]_1 \to \infty$ and by Lemma 5.1 we can conclude $z_{1,n}>0$.

We first consider the extreme case that $\hat{\rho}_1=\hat{\rho}_2=\infty$. Hence we have $u(\infty)=(1-\gamma_1)/\gamma_1$, $v(\infty)=\gamma_2/(1-\gamma_2)$, $\alpha_1=\log\{(1-\gamma_2)/\gamma_1\}$ and $\alpha_2=\log\{\gamma_2/(1-\gamma_1)\}$. By definition of $\Lambda$,
\begin{center}
	$\Lambda\alpha=b\pi_1\left(\begin{array}{cc}
	\rho_1 & 1  \\
	1 & \rho_2  \\
	\end{array}
	\right)\left(\begin{array}{cc}
	\gamma_1 & 1-\gamma_1  \\
	(1-\gamma_2)\tau & \gamma_2\tau  \\
	\end{array}
	\right)\left(\begin{array}{c}
	\log\frac{1-\gamma_2}{\gamma_1}  \\
	\log\frac{\gamma_2}{1-\gamma_1}  \\
	\end{array}
	\right)
	$
\end{center}

\begin{center}
	$=b\pi_1\left(\begin{array}{cc}
	\rho_1 & \tau  \\
	1 & \rho_2\tau  \\
	\end{array}
	\right)\left(\begin{array}{cc}
	\gamma_1 & 1-\gamma_1  \\
	(1-\gamma_2) & \gamma_2  \\
	\end{array}
	\right)\left(\begin{array}{c}
	\log\frac{1-\gamma_2}{\gamma_1}  \\
	\log\frac{\gamma_2}{1-\gamma_1}  \\
	\end{array}
	\right)$
\end{center}

\begin{center}
	$=b\pi_1\left(\begin{array}{cc}
	\rho_1 & \tau  \\
	1 & \rho_2\tau  \\
	\end{array}
	\right)\left(\begin{array}{c}
	-D(\gamma_1||(1-\gamma_2))  \\
	D((1-\gamma_2)||\gamma_1)  \\
	\end{array}
	\right)$.
\end{center}

So $[\Lambda\alpha]_1$ has the form

\begin{align}
[\Lambda \alpha]_1=b [ \pi_1\{\tau D((1-\gamma_2)||\gamma_1)-\rho_1 D(\gamma_1||(1-\gamma_2))\} ], \nonumber
\end{align}¡£

Since $\gamma_1,\gamma_2 \neq 1/2$ and
\begin{align}
\frac{\tau}{\rho_1}(1+\delta) \leq \frac{D(\gamma_1||(1-\gamma_2))) }{D((1-\gamma_2)||\gamma_1)} \leq (1-\delta) \rho_2 \tau,\nonumber
\end{align}
it is easy to see that $[\Lambda \alpha]_1<0$ when $\hat{\rho}_1=\hat{\rho}_2=\infty$. And therefore it is also true for sufficiently large $\hat{\rho}_1$ and $\hat{\rho}_2$. Moreover,   $[\Lambda\alpha]_1 \rightarrow -\infty$ when $b\rightarrow \infty$. And we can have similar result of  $[\Lambda\alpha]_2 \to \infty$.

In addition, for sufficiently large value of $n$, $[\Lambda \alpha]_1 \le 3\norm{\alpha}_{\infty}[\Lambda|\alpha|]_1$.

Putting pieces together, we have
\begin{align}\label{pr1}
\textnormal{pr}[\xi_i(\sigma)\ge -\log \hat{\tau}_{(1)}] \le \exp \left\{  -\frac{z_{1,n}^2}{4\norm{\alpha}_{\infty} (\Lambda|\alpha|)_1}   \right\}.
\end{align}
Pick $u_n^1$ satisfying $$u_n^1\log u_n^1=\frac{2C}{e\pi_1\bar{p}_1\{\log \hat{\tau}_{(1)}\}}, $$
where $\bar{p}_1\{\log \hat{\tau}_{(1)}\}=\frac{1}{n_1}\sum \limits_{i=1}^{n_1}\textnormal{pr}[\xi_i(\sigma)\ge -\hat{\tau}_{(1)}]$. We have

\begin{equation}
\textnormal{pr}[\sup \limits_{e \in \mathcal{E}}M_{n,1}(\V{e}) > \frac{1}{\pi_1}\frac{2C}{\log u_n^1}] \le  \exp \{-n(C-r_n)\} , \nonumber
\end{equation}
by the same arguments in the supplement material of \cite{AAA}, where $C$ is a constant and $r_n=o(1/n)$.

The right hand side of \eqref{pr1} goes to 0 as $b \rightarrow \infty$. Therefore, $\log u_n^1 \rightarrow \infty$, which implies for any $\epsilon>0$,
\begin{equation}\label{111}
\textnormal{pr}[\sup \limits_{e \in \mathcal{E}}M_{n,1}(\V{e}) > \epsilon] \rightarrow 0, \mbox{ as } n\rightarrow \infty.
\end{equation}

%
By a similar argument as above, for $i \in \mathcal{C}_2$,
\begin{align}\label{222}
\textnormal{pr}[\sup \limits_{e \in \mathcal{E}}M_{n,2}(\V{e}) > \epsilon] \rightarrow 0, \mbox{ as } n \rightarrow \infty,
\end{align}
where ${M}_{n,2}(\V{e})  =(1/{n_2})\sum  \limits_{i \in \mathcal{C}_2} 1\{\hat{c}_i(\V{e})\neq 2\}$.
The result of the theorem will automatically follows by putting \eqref{111} and \eqref{222} together, i.e., $M_{n}(\V{e})=\pi_1M_{n,1}(\V{e})+\pi_2M_{n,2}(\V{e})$. This competes our proof to the theorem.



\end{document}